# On the Application of Data-Driven Deep Neural Networks in Linear and Nonlinear Structural Dynamics

Nan Feng[1], Guodong Zhang[2] and Kapil Khandelwal[3]

[1]Graduate student, Dept. of Civil & Env. Engg. & Earth Sci., University of Notre Dame
[2]Postdoctoral Associate, Dept. of Civil & Env. Engg. & Earth Sci., University of Notre Dame
[3]Associate Professor, Dept. of Civil & Env. Engg. & Earth Sci., 156 Fitzpatrick Hall, University of Notre Dame, Notre Dame, IN 46556, United States, Email: kapil.khandelwal@nd.edu
ORCID: 0000-0002-5748-6019, (Corresponding Author)

*Preprint Submitted*

## Abstract

The use of deep neural network (DNN) models as surrogates for linear and nonlinear structural dynamical systems is explored. The goal is to develop DNN based surrogates to predict structural response, i.e., displacements and accelerations, for given input (harmonic) excitations. In particular, the focus is on the development of efficient network architectures using fully-connected, sparsely-connected, and convolutional network layers, and on the corresponding training strategies that can provide a balance between the overall network complexity and prediction accuracy in the target dataspaces. For linear dynamics, sparsity patterns of the weight matrix in the network layers are used to construct convolutional DNNs with sparse layers. For nonlinear dynamics, it is shown that sparsity in network layers is lost, and efficient DNNs architectures with fully-connected and convolutional network layers are explored. A transfer learning strategy is also introduced to successfully train the proposed DNNs, and various loading factors that influence the network architectures are studied. It is shown that the proposed DNNs can be used as effective and accurate surrogates for predicting linear and nonlinear dynamical responses under harmonic loadings.

# 1  Introduction

In recent years, deep neural network (DNN) based models have received considerable attention due to the rapid development in computational hardware and accessibility of big data [1-4]. With these developments, DNN models are now widely used in applications ranging from image classification [5, 6], speech recognition [7, 8], machine translation systems [9, 10], natural language processing [11, 12], among others. In parallel, there is also an increasing interest in applying DNNs in structural analysis [13, 14] and design [15], e.g., utilization of DNN as surrogates for structural static response [16, 17], failure probability prediction [18, 19], damage detection [20-22], parameter estimation of hysteretic structural systems [23] and structural topology optimization [24-26], among others.

In *structural dynamics* applications, which is the focus of this paper, early attempts of using neural network models can be traced back to the work by Masri et al. [27] where a feedforward neural network (FFNN) was trained to predict the restoring force given structural displacement and velocity at specified discrete-time instants for single-degree-of-freedom (SDOF) systems. In their study, an FFNN architecture with two hidden layers was chosen and network training was performed via a back-propagation algorithm [28]. With the advancements in learning algorithms for training deep network architectures and the availability of graphical processing units (GPUs) for data processing on large datasets [2], there is a renewed interest in the application of neural networks in structural dynamics applications. For instance, Wu et al. [29] used neural networks to predict the response of dynamical systems under earthquake loading using FFNN and convolutional neural network (CNN) models. Zhang et al. [30] applied long short-term memory (LSTM) recurrent networks to predict the structural response of a nonlinear multiple-degree-of-freedom (MDOF) system under seismic excitation. Stoffel et al. [31, 32] carried out a comparative study on the performance of different types of neural networks for approximating the deflection of an inelastic plate under an out-of-plane impulse load. Moreover, Yu et al. [33] proposed a residual recurrent neural network architecture to predict the structural responses of linear and nonlinear systems under sinusoidal or earthquake excitations. In the framework of so-called physics-informed DNNs, Zhang et al. [34] proposed to use a combination of both the standard mean squared error (MSE) of the output data and a physics loss term to train a CNN model taking ground motion acceleration as input and structural response variables as output. The physics loss term consists of an MSE of velocity computed via CNN prediction and finite difference based on the predicted displacement, and the MSE of a constraint that forces CNN outputs to satisfy the underlying governing equations. The authors later [35] used a similar loss function to train several self-tailored LSTM networks for predicting displacement, velocity, and restoring force of nonlinear structures considering a hysteresis model. In another study, Eshkevari et al. [36] designed a recurrent neural network to predict the response quantities including displacement, velocity, internal force, and acceleration for nonlinear systems subjected to ground motion acceleration.

Previous studies have shown the potential of DNNs in structural dynamics applications and many neural networks and training strategies have been presented to solve various problems. However, this is still a nascent field, and many issues have to be addressed before these networks can be successfully used in applications. In particular, in the context of SDOF and MDOF dynamical systems, two related issues must be addressed: a) design of effective network architectures, and b) construction and specification of dataspaces in which the network models are valid. For instance, in many cases, the networks are designed on an ad-hoc basis with many network layers and there is a lack of a systematic exploration of network architectures. With this architecture design strategy, efficient architectures with fewer training parameters may be overlooked. This is an important issue, as it is possible to choose a complex network architecture for a simple target system, which even with regularization is more likely to lead to overfitting [37]. Moreover, there is an increased computational overhead with large networks, as storage, training, and evaluation of large networks is expensive due to the involved floating-point operations. To address this issue, network architectures should be determined in accordance with a balance between model complexity and prediction accuracy. More importantly, when neural network models are used as data-driven surrogates, their validity is only guaranteed within the dataspaces in which they are designed and trained. For instance, a larger target design dataspace might require a deeper network as compared to a smaller dataspace. However, the characterization of the underlying modeling dataspaces is often missing, and therefore, the range of applicability of the trained networks is not clear.

In this exploratory study, DNN models for linear and nonlinear structural dynamics analysis are systematically investigated. The focus is on the construction of efficient network architectures using fully connected, sparsely connected, and convolutional network layers, and on the corresponding training strategies that can provide a balance between the overall network complexity and prediction accuracy. Considering a linear combination of harmonic functions, the input dataspace for a DNN model construction and training is explicitly expressed within which the trained DNN model can be applied. In particular, for linear structural dynamic responses, sparsity patterns of the fully connected layers in the network models are discovered. These sparsity patterns are then used to construct sparsely connected layers to reduce the trainable model parameters, and to compensate for the sparsity parameter-efficient convolutional layers are introduced. For nonlinear dynamics, it is shown that sparsity in network layers is lost, and efficient DNNs architectures with fully connected and convolutional network layers are explored. A transfer learning strategy is also presented to train the proposed DNNs, and various factors that influence the network architectures such as the duration, frequency range, number of frequencies in the load, and nonlinearity of the structure are studied. The rest of the paper is organized as follows: Section 2 describes the structural dynamics problems considered in this study. Section 3 presents the network architecture and training strategies. In Section 4, the proposed network architectures and training methodology are applied to linear and nonlinear SDOF dynamical systems considering harmonic loading with different parameter ranges. In Section 5, the application of DNNs is extended to linear and nonlinear MDOF systems. Finally, conclusions and important remarks are given in Section 6.

## 2    Dynamical Systems Description

In this study, neural networks (NNs) are used to predict the response of a single degree of freedom (SDOF) and multiple degrees of freedom (MDOF) dynamical systems subjected to harmonic loadings that are randomly sampled from prescribed dataspaces. Here, appropriate dataspaces are chosen according to the available hardware and computing resources, as larger dataspaces require more memory and training time to train the networks to reach a satisfactory performance. These dynamical systems and the corresponding dataspaces are described in the following sections.

### 2.1  SDOF system

#### 2.1.1  Linear case

For a linear SDOF system, the equation of motion reads

$$m\ddot{u} + c\dot{u} + ku = p(t) \tag{1}$$

where $m$, $k$ and $c$ are mass, stiffness and damping coefficient, respectively; $u = u(t)$ is displacement dependent on time $t$ and $p(t)$ is the applied load. The following parameters are used to model the SDOF system: mass $m$ = 13.5 kg, fundamental period of system $T_n$ = 0.5 sec and natural frequency of the system $\omega_n = 2\pi/T_n$ = 12.6 rad/sec, and the damping ratio is taken as $\xi$ = 0.02. Thus, the stiffness is $k = m\omega_n^2$ = 2131.8 kg/sec$^2$ and the damping coefficient is $c = 2m\omega_n\xi$ = 6.8 kg/sec. The initial conditions are $u(0) = 0$ and $\dot{u}(0) = 0$.

The input to this system is the loading $p(t)$, and the response quantities of interest are the displacement $u(t)$ and acceleration $\ddot{u}(t)$ within a prescribed time window $t \in [0, T]$. The harmonic loading $p(t)$ is modeled as a linear combination of harmonic components and is given by

$$p(t) = \sum_{i=1}^{N_p} a_i \sin(\omega_i t + \phi_i) \tag{2}$$

where $N_p$ is the number of harmonic terms considered; $a_i$, $\omega_i$ and $\phi_i$ are the amplitudes, frequencies, and phase angles in the harmonic terms ($i = 1, 2, ..., N_p$), respectively, that are modeled as uniformly distributed random variables with probability densities given by $a_i \sim \mathcal{U}(0,10)$, $\omega_i \sim \mathcal{U}(0,6\pi)$ and $\phi_i \sim \mathcal{U}(0,2\pi)$. Thus, the underlying input dataspace can be described by $\mathcal{D}_{\text{lsdof}} = \{a_i \sim \mathcal{U}(0,10), \varphi_i \sim \mathcal{U}(0,2\pi), \text{ and } \omega_i \sim \mathcal{U}(0,6\pi) \text{ with } i = 1,2,\cdots,N_p\}$.

The response quantities at uniformly spaced discrete times ($t_k$) are of interest, where $t_k = k\Delta t$ ($k = 0, 1, 2 \ldots, n_s$) with $\Delta t = T/n_s$ and $n_s$ is the total number of time steps. Responses are represented by $n \times 1$ vectors as $\boldsymbol{y} = [u(t_0), u(t_1), ..., u(t_{n_s})]$ or $\boldsymbol{y} = [\ddot{u}(t_0), \ddot{u}(t_1), ..., \ddot{u}(t_{n_s})]$, where $n = n_s + 1$. For a given input $p(t) \in \mathcal{D}_{\text{lsdof}}$, the response $\boldsymbol{y}$ of the dynamical system in Eq. (1) can be computed using analytical solutions provided in Appendix A.1.

### 2.1.2   Nonlinear case

For a nonlinear SDOF system with a cubic spring, the equation of motion reads

$$m\ddot{u} + c\dot{u} + ku + bu^3 = p(t) \tag{3}$$

where $m, c, k$ are taken the same values as those in Section 2.1.1. The coefficient in the cubic term $b$ is chosen from the set $\{0.25k, 0.5k, 0.75k, k\}$.

The underlying input dataspace for the harmonic loading $p(t)$ in Eq. (2) is considered to be $\mathcal{D}_{\text{nlsdof}} = \{a_i \sim \mathcal{U}(0,20), \varphi_i \sim \mathcal{U}(0,2\pi), \text{ and } \omega_i \sim \mathcal{U}(0,6\pi) \text{ with } i = 1,2,\cdots,N_p\}$. The response quantities of interest in this case are still displacement or acceleration at discrete times and the response $\boldsymbol{y}$ of the dynamical system in Eq. (3) is computed using Newmark-$\beta$ method [38].

## 2.2   MDOF system

### 2.2.1   Linear case

For the linear elastic MDOF case, a dynamical system with six degrees of freedom is considered. This system is used to model a six-story shear frame with Rayleigh damping and is governed by

$$\boldsymbol{M}\ddot{\boldsymbol{u}} + \boldsymbol{C}\dot{\boldsymbol{u}} + \boldsymbol{K}\boldsymbol{u} = \boldsymbol{p}(t) \tag{4}$$

where $\boldsymbol{u} = \{u_i(t)\}_{i=1}^6$ represents displacement vector and $\dot{\boldsymbol{u}}$ and $\ddot{\boldsymbol{u}}$ are the 1st-order and 2nd-order time derivative of $\boldsymbol{u}$ representing velocity and acceleration vectors, respectively, and

$$\boldsymbol{M} = \begin{bmatrix} m_1 & 0 & 0 & 0 & 0 & 0 \\ 0 & m_1 & 0 & 0 & 0 & 0 \\ 0 & 0 & m_2 & 0 & 0 & 0 \\ 0 & 0 & 0 & m_2 & 0 & 0 \\ 0 & 0 & 0 & 0 & m_3 & 0 \\ 0 & 0 & 0 & 0 & 0 & m_3 \end{bmatrix}$$

$$\boldsymbol{K} = \frac{EI}{H^3} \begin{bmatrix} 48 & -24 & 0 & 0 & 0 & 0 \\ -24 & 48 & -24 & 0 & 0 & 0 \\ 0 & -24 & 48 & -24 & 0 & 0 \\ 0 & 0 & -24 & 48 & -24 & 0 \\ 0 & 0 & 0 & -24 & 48 & -24 \\ 0 & 0 & 0 & 0 & -24 & 24 \end{bmatrix} \qquad (5)$$

$$\boldsymbol{C} = \alpha_0 \boldsymbol{M} + \alpha_1 \boldsymbol{K}$$

$$\alpha_0 = \frac{2\xi \omega_{n1} \omega_{n2}}{\omega_{n1} + \omega_{n2}} \text{ and } \alpha_1 = \frac{2\xi}{\omega_{n1} + \omega_{n2}}$$

$$\boldsymbol{p}(t) = -a_g(t)\boldsymbol{M}[1 \quad 1 \quad 1 \quad 1 \quad 1 \quad 1]^T$$

where $m_i$ ($i = 1,2,3$) are the story mass; $E$, $I$ and $H$ are Young's modulus, moment of inertia and story height, respectively; $\xi$ is the damping ratio for the system, and $\omega_{n1}$ and $\omega_{n2}$ are the first and second natural frequencies of the system, respectively. The values of these model parameters are summarized in Table 1.

In this case, the load is applied in terms of harmonic ground motion acceleration, $a_g(t)$, and is modeled as the sum of sinusoidal components as

$$a_g(t) = \sum_{i=1}^{N_p} a_i \sin(\omega_i t + \phi_i) \qquad (6)$$

where $a_i$, $\omega_i$ and $\phi_i$ are amplitudes, frequencies, and phases of the harmonic terms that follow uniform distributions, i.e., $a_i \sim \mathcal{U}(0,4)$, $\omega_i \sim \mathcal{U}(0,20\pi)$ and $\phi_i \sim \mathcal{U}(0,2\pi)$, respectively. The range of the harmonic frequencies is chosen such that the first two natural frequencies are covered. Thus, the underlying input dataspace can be described by $\mathcal{D}_{\text{lmdof}} = \{a_i \sim \mathcal{U}(0,4), \phi_i \sim \mathcal{U}(0,2\pi)$, and $\omega_i \sim \mathcal{U}(0,20\pi)$ with $i = 1,2,\cdots, N_p\}$.

In this case, the displacement and acceleration responses at the $6^{th}$ story at uniformly spaced discrete times are of interest, i.e. $\boldsymbol{y} = [u_6(t_0), u_6(t_1), \ldots, u_6(t_{n_s})]$ or $\boldsymbol{y} = [\ddot{u}_6(t_0), \ddot{u}_6(t_1), \ldots, \ddot{u}_6(t_{n_s})]$, which can be computed by using analytical solutions in Appendix A.2.

### 2.2.2 Nonlinear case

A nonlinear MDOF system with cubic springs is considered, whose response is governed by

$$\boldsymbol{M}\ddot{\boldsymbol{u}} + \boldsymbol{C}\dot{\boldsymbol{u}} + \boldsymbol{F}_{int}(\boldsymbol{u}) = \boldsymbol{p}(t) \tag{7}$$

where the mass matrix $\boldsymbol{M}$ and damping matrix $\boldsymbol{C}$ keep the same as those in Section 2.2.1. The internal force is given by

$$\boldsymbol{F}_{int}(\boldsymbol{u}) = \begin{bmatrix} (ku_1 + bu_1^3) + k(u_1 - u_2) + b(u_1 - u_2)^3 \\ k(u_2 - u_1) + b(u_2 - u_1)^3 + k(u_2 - u_3) + b(u_2 - u_3)^3 \\ k(u_3 - u_2) + b(u_3 - u_2)^3 + k(u_3 - u_4) + b(u_3 - u_4)^3 \\ k(u_4 - u_3) + b(u_4 - u_3)^3 + k(u_4 - u_5) + b(u_4 - u_5)^3 \\ k(u_5 - u_4) + b(u_5 - u_4)^3 + k(u_5 - u_6) + b(u_5 - u_6)^3 \\ k(u_6 - u_5) + b(u_6 - u_5)^3 \end{bmatrix} \tag{8}$$

where the stiffness coefficient is $k = 24EI/H^3$ and the value of the cubic coefficient $b$ is chosen from the set $\{0.25k, 0.5k, 0.75k, k\}$. It is remarked that the nonlinear system in Eq. (7) will degenerate to the linear system in Eq. (4) if the cubic coefficient is chosen as $b = 0$.

In this case, the load $\boldsymbol{p}(t)$ is again applied in terms of harmonic ground motion acceleration $a_g(t)$. The underlying input dataspace for the ground motion acceleration $a_g(t)$ in Eq. (6) is considered to be $\mathcal{D}_{\text{nlmdof}} = \{a_i \sim \mathcal{U}(0,4), \varphi_i \sim \mathcal{U}(0,2\pi), \text{ and } \omega_i \sim \mathcal{U}(0,10\pi) \text{ with } i = 1,2,\cdots,N_p\}$. The same response quantities in Section 2.2.2 are of interest and are computed using Newmark-$\beta$ method [38].

## 3 Neural Network Design and Training

In this section, NNs are designed and trained such that they can be used as surrogates to the dynamical systems considered in Section 2. Neural networks use compositions of relatively simple mathematical functions to represent complex mappings. In particular, in many NNs these mathematical operations are carried out in a layer-by-layer fashion [2], where an output of one layer is taken as an input to the next layer. The design of NNs involves determining the architecture of each layer and the number of such data processing layers in the network while training a NN involves determining the network parameters ($\boldsymbol{\theta}$) that are used to describe these computational layers.

For the considered dynamical systems, the input of the NN is the loading history given by an $n \times 1$ vector, i.e. $\boldsymbol{x} = \left[p(t_0), p(t_1), \ldots, p(t_{n_s})\right]$ or $\boldsymbol{x} = \left[a_g(t_0), a_g(t_1), \ldots, a_g(t_{n_s})\right]$, which is sampled from the underlying dataspace $\mathcal{D}$, where $\mathcal{D}$ is taken to be $\mathcal{D}_{\text{lsdof}}$ or $\mathcal{D}_{\text{nlsdof}}$ for linear or nonlinear SDOF problems, and $\mathcal{D}_{\text{lmdof}}$ or $\mathcal{D}_{\text{nlmdof}}$ for linear or nonlinear MDOF problems. The output of the NN is the response of the system that includes displacement and acceleration time history, e.g., $\boldsymbol{y} = \left[u(0), u(1), \ldots, u(t_{n_s})\right]$ for the time series of displacement of SDOF. In essence, NN represents a mapping $\boldsymbol{f}_{nn}$ parameterized by $\boldsymbol{\theta}$ that maps the input to the output, i.e., $\boldsymbol{f}_{nn}: \boldsymbol{x} \mapsto \boldsymbol{y}$ where $\boldsymbol{x}, \boldsymbol{y} \in \mathbb{R}^n$. Thus, the task of a network is to predict $\boldsymbol{y}$ from $\boldsymbol{x}$ by outputting $\hat{\boldsymbol{y}} = \boldsymbol{f}_{nn}(\boldsymbol{x}; \boldsymbol{\theta})$ and the goal of machine learning is to learn the parameters $\boldsymbol{\theta}$.

From a probabilistic regression viewpoint [39], the predictive conditional PDF $p(\boldsymbol{y}|\boldsymbol{x})$ reads

$$p(\boldsymbol{y}|\boldsymbol{x}) = \mathcal{N}(\boldsymbol{y}|\boldsymbol{f}_{nn}(\boldsymbol{x}; \boldsymbol{\theta}), \sigma^2 \boldsymbol{I}) \tag{9}$$

In addition, by assuming that all training data pairs $(\boldsymbol{x}_i, \boldsymbol{y}_i)$ $(i = 1, 2, \cdots, N_{train})$ are i.i.d., the joint PDF of all training data points read

$$p(\boldsymbol{y}_1, \boldsymbol{y}_2, \cdots, \boldsymbol{y}_{N_{train}} | \boldsymbol{x}_1, \boldsymbol{x}_2, \cdots, \boldsymbol{x}_{N_{train}}) = \prod_{i=1}^{N_{train}} p(\boldsymbol{y}_i | \boldsymbol{x}_i) \tag{10}$$

The joint PDF in Eq. (10) can be considered as a measure for quantifying how well the proposed NN model fits the observed training data points. Thus, the value of network parameters $\boldsymbol{\theta}$ can be obtained by solving the following maximum likelihood estimation (MLE) optimization problem

$$\boldsymbol{\theta}^* = \underset{\boldsymbol{\theta}}{\mathrm{argmax}} \ln p(\boldsymbol{y}_1, \boldsymbol{y}_2, \cdots, \boldsymbol{y}_{N_{train}} | \boldsymbol{x}_1, \boldsymbol{x}_2, \cdots, \boldsymbol{x}_{N_{train}}) \tag{11}$$

Finally, the learned conditional distribution $p(\boldsymbol{y}|\boldsymbol{x})$ is $\mathcal{N}(\boldsymbol{y}|\boldsymbol{f}_{nn}(\boldsymbol{x}; \boldsymbol{\theta}^*), \sigma^2 \boldsymbol{I})$ and the most probable NN mapping is given by $\hat{\boldsymbol{y}} = \boldsymbol{f}_{nn}(\boldsymbol{x}; \boldsymbol{\theta}^*)$.

### 3.1 Fully-connected (FC) neural network

The neural network architectures used to model dynamical systems in this study are shown in Figure 1a. The network starts with an input layer $\boldsymbol{x}$ of size $n \times 1$ representing the external loading time sequence. The output layer $\boldsymbol{y}$, i.e., displacement or acceleration, has the same size as the input layer. For a standard ($l^{th}$) FC layer (Figure 1a), the weighted sum operation is applied to the input vector $\boldsymbol{x}^{[l-1]}$ from the $(l-1)^{th}$ layer using a $n_H^{[l]} \times n_H^{[l-1]}$ weight matrix $\boldsymbol{W}^{[l]}$ and $n_H^{[l]} \times 1$ bias vector $\boldsymbol{b}^{[l]}$, where $n_H^{[l-1]}$ and $n_H^{[l]}$ represent the size of the $(l-1)^{th}$ and $l^{th}$ layers, respectively. This operation gives

$$\boldsymbol{h}_1^{[l]} = \boldsymbol{W}^{[l]} \boldsymbol{x}^{[l-1]} + \boldsymbol{b}^{[l]} \tag{12}$$

where $\boldsymbol{h}_1^{[l]}$ is a $n_H^{[l]} \times 1$ vector.

After the weighted sum in Eq. (12), a linear (identity) activation or a nonlinear ReLU activation is applied depending on whether the FC layer is the last layer of the network or not. Thus, the output vector of the $l^{th}$ layer reads

$$\boldsymbol{x}^{[l]} = \begin{cases} \boldsymbol{h}_1^{[l]} & \text{if layer } l \text{ is the last layer} \\ \Phi\left(\boldsymbol{h}_1^{[l]}\right) & \text{otherwise} \end{cases} \qquad (13)$$

where $\boldsymbol{x}^{[l]}$ is a $n_H^{[l]} \times 1$ vector and $\Phi(\blacksquare)$ represents an element-wise ReLU activation function, i.e.,

$$\Phi(x) = \begin{cases} x & \text{if } x > 0 \\ 0 & \text{otherwise} \end{cases} \qquad (14)$$

### 3.2   Sparse convolutional neural networks

In some cases, the trained FC layers in Section 3.1 exhibit a sparsity pattern, i.e., there are many small weights in the layers. Although these small weights help to improve the model performance, they also lead to increased computational cost in terms of both network evaluation and parameter optimization. To make use of the underlying sparsity pattern while keeping the desired network accuracy the convolutional layer enhanced sparse networks are proposed in this study. The overall idea of these sparse networks is to use the sparsity pattern to construct sparse FC layers that are then used together with convolutional layers. The added convolutional layers provide additional capacity that compensates for the sparsity of the FC layer. As the convolutions layers provide parameter sharing, they do not significantly increase the number of trainable parameters.

In Figure 1b, the sparse network architecture contains, between the input and output layers, convolutional (CONV) layers, batch normalization (BN) layers, and sparsely connected (SC) layers. BN layers are added before each CONV layer, except the first CONV layer which is preceded by the input layer that has already been normalized (Section 3.3.1). In the cases where higher representational capacity is required, more batch-convolutional blocks (BN-CONV) that comprise a batch normalization and a convolutional layer are added. In general, increasing the number of BN-CONV layers will increase the overall representational capacity of the network. However, this will also lead to a deeper NN model with more training parameters ($\boldsymbol{\Theta}$), and thus may require more sophisticated training strategies. The computational operations in each of these layers are described in the following sections.

### 3.2.1 Sparsely connected (SC) layer

To construct an SC layer, the NN model with FC layers is first trained. Next, a *sparsity analysis* on the trained weight matrices $\boldsymbol{W}^{[l]}$ ($l \rightarrow$ index of hidden layers) is carried out with a predefined threshold value such that the weights that are smaller than the threshold are taken as zero to determine the sparsity pattern of weight matrices. The zero weights in the resulting sparse weight matrices are then fixed and are not updated during the training process. As an illustration, Figure 2 shows a weight matrix of one FC layer where the intensity of color denotes the relative magnitudes of the weights. With a pre-defined threshold, only the weights that are larger than the threshold are kept for training, and this results in a sparse representation shown in Figure 2.

### 3.2.2 Convolutional (CONV) layer

In a CONV layer, padding, convolution, and activation are the main operations that are used to extract data features and add nonlinearities to the architecture. Suppose the $l^{\text{th}}$ layer is a CONV layer, padding, convolution, and activation are applied sequentially to the data from the previous layer $\boldsymbol{x}^{[l-1]}$ of dimension $n_H^{[l-1]} \times n_C^{[l-1]}$. For padding with dimension $p^{[l]}$, a volume of dimension $p^{[l]} \times n_C^{[l-1]}$ containing only zeros is concatenated on two ends of the volume $\boldsymbol{x}^{[l-1]}$ and the output from this operation is $\boldsymbol{h}_1^{[l]} = [\boldsymbol{0}; \boldsymbol{x}^{[l-1]}; \boldsymbol{0}]$. The size of the volume $\boldsymbol{h}_1^{[l]}$ is then $(2p^{[l]} + n_H^{[l-1]}) \times$

$n_C^{[l-1]}$. After padding, convolution operation is applied to volume $\boldsymbol{h}_1^{[l]}$ by using $n_C^{[l]}$ filters. The

size of each filter is $f^{[l]} \times n_C^{[l-1]}$ and the filter stride used is denoted by $s^{[l]}$. In each filter, there are weights $\boldsymbol{W}_i^{[l]}$ of size $f^{[l]} \times n_C^{[l-1]}$ and scalar biases $b_i^{[l]}$ $(i = 1, 2, \cdots, n_C^{[l]})$. Using the $i^{\text{th}}$ filter,

the convolution operation outputs a vector:

$$\boldsymbol{v}_i = \boldsymbol{h}_1^{[l]} * \boldsymbol{W}_i^{[l]} = \begin{bmatrix} \boldsymbol{h}_1^{[l]}[1:f^{[l]}] \star \boldsymbol{W}_i^{[l]} + b_i^{[l]} \\ \boldsymbol{h}_1^{[l]}[s^{[l]}+1:s^{[l]}+f^{[l]}] \star \boldsymbol{W}_i^{[l]} + b_i^{[l]} \\ \boldsymbol{h}_1^{[l]}[2s^{[l]}+1:2s^{[l]}+f^{[l]}] \star \boldsymbol{W}_i^{[l]} + b_i^{[l]} \\ \vdots \\ \boldsymbol{h}_1^{[l]}[2p^{[l]}+n_H^{[l-1]}-f^{[l]}+1:2p^{[l]}+n_H^{[l-1]}] \star \boldsymbol{W}_i^{[l]} + b_i^{[l]} \end{bmatrix} \qquad (15)$$

$i = 1, 2, \cdots, n_C^{[l]}$

with dimension

$$\left( \frac{2p^{[l]} + n_H^{[l-1]} - f^{[l]}}{s^{[l]}} + 1 \right) \times 1 \qquad (16)$$

and for two volumes $\boldsymbol{a}$ and $\boldsymbol{b}$ of the same size $n_1 \times n_2$, the operation $\star$ is given by

$$\boldsymbol{a} \star \boldsymbol{b} \triangleq \sum_{i=1}^{n_1} \sum_{j=1}^{n_2} \boldsymbol{a}[i,j]\boldsymbol{b}[i,j] \qquad (17)$$

In Eq. (15), $\boldsymbol{h}_1^{[l]}[p:q]$ means that the volume $\boldsymbol{h}_1^{[l]}$ is sliced along the height from $p^{th}$ to $q^{th}$ rows.

It is noted that the padding size $p^{[l]}$ and filter stride $s^{[l]}$ are often chosen such that the size in Eq.

(16) is an integer.

Finally, the vectors $\boldsymbol{v}_i$ $(i = 1, 2, \cdots, n_C^{[l]})$ are stacked together to output a new volume

$$\boldsymbol{h}_2^{[l]} = \begin{bmatrix} | & | & & | \\ \boldsymbol{v}_1 & \boldsymbol{v}_2 & ... & \boldsymbol{v}_{n_C^{[l]}} \\ | & | & & | \end{bmatrix}_{n_H^{[l]} \times n_C^{[l]}} \tag{18}$$

where

$$n_H^{[l]} = \frac{2p^{[l]} + n_H^{[l-1]} - f^{[l]}}{s^{[l]}} + 1 \tag{19}$$

In this study, it is required that the output volume $\boldsymbol{h}_2^{[l]}$ has the same height as $\boldsymbol{x}^{[l-1]}$ such that the convolutional layer does not change the height of the input volume. Hence, the padding size is chosen to be

$$p^{[l]} = \frac{1}{2}\left[\left(n_H^{[l-1]} - 1\right)s^{[l]} + f^{[l]} - n_H^{[l-1]}\right] \tag{20}$$

With this padding, $n_H^{[l]} = n_H^{[l-1]}$ and the size of the volume $\boldsymbol{h}_2^{[l]}$ becomes $n_H^{[l-1]} \times n_C^{[l]}$.

To add nonlinearities in the layer, the rectified linear unit (ReLU) [1] is used as an activation function and is applied to each element in the volume $\boldsymbol{h}_2^{[l]}$. As a result, the output volume of this convolutional layer is

$$\boldsymbol{x}^{[l]} = \Phi\left(\boldsymbol{h}_2^{[l]}\right) \tag{21}$$

where the activation ($\Phi$) is an element-wise operation and does not change the size of the volume $\boldsymbol{h}_2^{[l]}$. Depending on the location of the CONV layer, linear or ReLU activation is applied. In particular, linear activation is used for the last CONV layer whose functionality is mainly for resizing the data vector to $n_H^{[0]} \times 1$, while the ReLU activation function is applied to all the other middle CONV layers.

### 3.2.3 Batch normalization (BN) layer

The BN layer is designed to address the so-called *internal covariate drift* issue, which refers to a phenomenon that in the training process the distribution of each layer's input $\boldsymbol{x}^{[l-1]}$ continue to change, as previous CONV or FC layer's weights $\boldsymbol{W}^{[l-1]}$ and biases $\boldsymbol{b}^{[l-1]}$ get updated. This phenomenon is found to significantly slow down the training [40]. With the BN layer, each input is normalized independently based on the statistics of the samples in the current mini-batch (explanation of mini-batch can be found in Section 3.3.3) [40]. This normalization helps to maintain a stable training process.

Consider that the $l^{\text{th}}$ layer in the architecture is a BN layer, the input volume $\boldsymbol{x}^{[l-1]}$ is normalized by its expectation and variance. The normalization operation is applied to each element of volume $\boldsymbol{x}^{[l-1]}$ and outputs a new volume $\boldsymbol{x}^{[l]}$ with the same size as $\boldsymbol{x}^{[l-1]}$, i.e., $n_H^{[l]} = n_H^{[l-1]}$ and $n_C^{[l]} = n_C^{[l-1]}$. The calculation of $\boldsymbol{x}^{[l]}$ given $\boldsymbol{x}^{[l-1]}$ is

$$\boldsymbol{x}^{[l]}[i,j] = \boldsymbol{\gamma}^{[l]}[i,j] \frac{\boldsymbol{x}^{[l-1]}[i,j] - E\left[\boldsymbol{x}^{[l-1]}[i,j]\right]}{\sqrt{Var\left[\boldsymbol{x}^{[l-1]}[i,j]\right] + \epsilon_{BN}}} + \boldsymbol{\beta}^{[l]}[i,j]$$

(22)

with $i = 1,2,\cdots,n_H^{[l-1]}$ and $j = 1,2,\cdots,n_C^{[l-1]}$

where $\boldsymbol{x}^{[l]}[i,j]$ denotes the element of volume $\boldsymbol{x}^{[l]}$ with index $[i,j]$; $\boldsymbol{\gamma}^{[l]}$ and $\boldsymbol{\beta}^{[l]}$ are matrices of size $n_H^{[l-1]} \times n_C^{[l-1]}$ that contains all the scaling and shifting parameters; $\epsilon_{BN}$ is a hyperparameter used to avoid numerical issues when the variance is close to zero. The shifting and scaling parameters ensure that the transformation in Eq. (22) can represent an identity [40], which means no operation may be done in the BN layer. Assuming there are $N_s$ samples in the current mini-batch and a collection of these data for the $(l-1)^{th}$ layer, i.e., $\boldsymbol{x}^{[l-1]}$, $D_N = \left\{\boldsymbol{x}_1^{[l-1]}, \boldsymbol{x}_2^{[l-1]}, \cdots, \boldsymbol{x}_{N_s}^{[l-1]}\right\}$, the statistics in Eq. (22) are estimated by

$$E\left[\boldsymbol{x}^{[l-1]}[i,j]\right] \approx \frac{1}{N_s}\sum_{k=1}^{N_s}\boldsymbol{x}_k^{[l-1]}[i,j]$$

$$Var\left[\boldsymbol{x}^{[l-1]}[i,j]\right] \approx \frac{1}{N_s}\sum_{k=1}^{N_s}\left(\boldsymbol{x}_k^{[l-1]}[i,j] - E\left[\boldsymbol{x}^{[l-1]}[i,j]\right]\right)^2$$

(23)

where the square operation is applied in an elementwise fashion in Eq. (23). It should be noted that in the training stage where a mini-batch stochastic gradient algorithm is used, the statistical quantities in Eq. (23) are computed based on a mini-batch training dataset at the current iteration. However, in the testing stage, the statistics in Eq. (23) are computed based on the complete training dataset.

## 3.3   Neural networks training

The neural network models considered in this study are parameterized by parameters $\boldsymbol{\theta} = \{\boldsymbol{W}, \boldsymbol{b}, \boldsymbol{\gamma}, \boldsymbol{\beta}\}$, where $\boldsymbol{W}$ and $\boldsymbol{b}$ are weight matrices and bias vectors from all the CONV and SC layers; $\boldsymbol{\gamma}$ and $\boldsymbol{\beta}$ are scaling and shifting parameters from all the BN layers in the network. The neural networks training refers to the optimization process of finding the optimum $\boldsymbol{\theta}^*$ in Eq. (11).

### 3.3.1   Data preprocessing

Although theoretically not necessary, data (input and output pairs) preprocessing has become a routine step in deep learning, as it can speed up learning and lead to better model performance. There are multiple approaches for data preprocessing [41, 42], and the commonly used normalization strategy is chosen in this study. For each component $\eta$ of the input as well as output, all the samples in the training dataset ($\eta_i$, $i = 1, \ldots, N_{train}$) are normalized by

$$\tilde{\eta}_i = \frac{\eta_i - \mu_\eta}{\sqrt{V_\eta + \epsilon_D}} \ , \quad i = 1, 2, \cdots, N_{\text{train}} \tag{24}$$

with

$$\mu_\eta = \frac{1}{N_{\text{train}}} \sum_{i=1}^{N_{\text{train}}} \eta_i \quad \text{and} \quad V_\eta = \frac{1}{N_{\text{train}}} \sum_{i=1}^{N_{\text{train}}} \left( \eta_i - \mu_\eta \right)^2 \tag{25}$$

where $\epsilon_D$ is added to avoid numerical issues when the value of variance is close to zero. This happens to the first component of output, as all samples follow the same initial condition, i.e., $u(t_0) = 0$. It should be noted that the testing dataset is normalized by the same statistics of the *training* dataset, i.e., $\mu_\eta$ and $V_\eta$ in Eq. (25), before using them as inputs to a network model. The final output is then obtained by carrying out the inverse of the transformation in Eq. (24) in a component-wise manner.

### 3.3.2   Loss function and error metric

As has been discussed at the beginning of Section 3, from a probabilistic viewpoint, the network parameters $\boldsymbol{\theta}$ can be determined by solving the maximum likelihood estimation (MLE) optimization problem in Eq. (11). Replacing the input/output pairs $(\boldsymbol{x}_i, \boldsymbol{y}_i)$ with the normalized one $(\tilde{\boldsymbol{x}}_i, \tilde{\boldsymbol{y}}_i)$ in Eqns. (9)-(11), the training task can be completed equivalently by finding optimal $\boldsymbol{\theta}$ in $\mathbb{R}^{n_p}$ to minimize the negative log-likelihood function, where $n_p$ is the number of model parameters, i.e.,

$$L_1(\boldsymbol{\theta}) = - \sum_{i=1}^{N_{train}} \ln p(\tilde{\boldsymbol{y}}_i | \tilde{\boldsymbol{x}}_i, \boldsymbol{\theta}) = \frac{1}{2\sigma_1^2} \sum_{i=1}^{N_{\text{train}}} \| \boldsymbol{f}_{nn}(\tilde{\boldsymbol{x}}_i; \boldsymbol{\theta}) - \tilde{\boldsymbol{y}}_i \|_2^2 + \frac{n N_{\text{train}}}{2} \ln 2\pi\sigma^2 \tag{26}$$

where $\| \blacksquare \|_2$ is the Euclidean norm of a vector. Thus, the training task can be further simplified by finding optimal $\boldsymbol{\theta}$ in $\mathbb{R}^{n_p}$ to minimize a squared error that reads

$$L_2(\boldsymbol{\theta}) = \frac{1}{N_{\text{train}}} \sum_{i=1}^{N_{\text{train}}} \|\boldsymbol{f}_{nn}(\widetilde{\boldsymbol{x}}_i; \boldsymbol{\theta}) - \widetilde{\boldsymbol{y}}_i\|_2^2 \tag{27}$$

Finally, when regularization is also considered, loss function in the learning process in this study is given by

$$\mathcal{L}(\boldsymbol{\theta}) = L_2(\boldsymbol{\theta}) + \lambda(\|\boldsymbol{W}\|_F^2 + \|\boldsymbol{b}\|_2^2) \tag{28}$$

where the first term in the loss function represents the application of likelihood principle over the training dataset ($D_{\text{train}}$) and the second term is used to regularize the loss function, in which $\lambda$ is a scalar regularization hyperparameter and $\|\blacksquare\|_F$ is the Frobenius matrix norm. The second term in the loss function is needed since, without this regularization term, the ML algorithm will always prefer to train a network model that minimizes the observation error over the training dataset regardless of the complexity of the trained model and chances are high that such models do not have good performance over the testing dataset, which is usually referred to as *overfitting*. By adding a regularization term in the loss function, the optimization algorithm will try to find a balance between the complexity of the trained model and the training error, which helps prevent overfitting issues. Apart from adding $L^2$ norm of parameters in the loss function, there are also other regularization options such as adding $L^1$ norm of parameters, early-stopping, random dropout, and model averaging [2].

Next, the unknown parameters in NN models are learned by solving an optimization problem, i.e.,

$$\boldsymbol{\theta}^* = \operatorname*{argmin}_{\boldsymbol{\theta} \in \mathbb{R}^{n_p}} \mathcal{L}(\boldsymbol{\theta}) \tag{29}$$

After training, for arbitrary observed input data point $\widetilde{\boldsymbol{x}}$, the predictive PDF of the output data $\widetilde{\boldsymbol{y}}$ reads

$$p(\widetilde{\boldsymbol{y}}|\widetilde{\boldsymbol{x}}, \boldsymbol{\theta}) = \mathcal{N}(\widetilde{\boldsymbol{y}}|\boldsymbol{f}_{nn}(\widetilde{\boldsymbol{x}}; \boldsymbol{\theta}^*), \sigma^2\boldsymbol{I}) \tag{30}$$

Therefore, from Eq. (30) the mean value of $\widetilde{\boldsymbol{y}}$ given by the NN model is $\widehat{\widetilde{\boldsymbol{y}}} = \boldsymbol{f}_{nn}(\widetilde{\boldsymbol{x}}; \boldsymbol{\theta}^*)$.

Finally, the performance ($\mathcal{P}$) of a NN model is measured in terms of the mean squared error of the model on the testing dataset using Eq. (27), i.e., $\mathcal{P}_{\text{mse}} \equiv L_2(\boldsymbol{\theta})$, with training data replaced with testing data. Note that a low value of $\mathcal{P}_{\text{mse}}$ indicates a good model performance, and vice versa. Another measure for the model prediction performance is through relative error metric

$$\varepsilon_r = \frac{\|\boldsymbol{y} - \widehat{\boldsymbol{y}}\|_2}{\|\boldsymbol{y}\|_2} \times 100\% \tag{31}$$

### 3.3.3   Stochastic gradient descent-based Adam optimizer

To solve the optimization problem in Eq. (29), stochastic gradient-descent-based learning algorithms are commonly employed [1], since calculating the true gradient over the entire training dataspace can slow down the training process and lead to prohibitively high memory requirement [2]. In particular, mini-batch stochastic gradient descent is used, where the training dataset $D_{\text{train}}$ is divided into $n_b$ mini-batches, i.e., $D_1, D_2, ..., D_{n_b}$. The number of samples in each batch $s_b = N_{\text{train}}/n_b$ is called the batch size. All mini-batch datasets are used one by one to compute a stochastic gradient for updating model parameters, and one cycle over all mini-batches is called an *epoch*. Hence, one epoch consists of $n_b$ optimization iterations. Since the gradient computed from the data in one mini-batch might be away from the true gradient over the entire training dataspace and might be very different from batch to batch, it can lead to an oscillating convergence behavior and slow down the training process. To address this issue, Adam optimizer considers moving statistics of the stochastic gradient, where at the $k^{\text{th}}$ optimization step in one epoch [43], the update process of an arbitrary trainable variable $\theta \in \boldsymbol{\theta}$ is given by

$$\theta_k = \theta_{k-1} - \alpha_k m_k \tag{32}$$

where

$$m_k = \frac{1}{1-(\beta_1)^k}[\beta_1 m_{k-1} + (1-\beta_1)g_k] \; , \quad \alpha_k = \frac{\alpha}{\sqrt{v_k} + \epsilon_l}$$

$$\text{with} \quad v_k = \frac{1}{1-(\beta_2)^k}[\beta_2 v_{k-1} + (1-\beta_2)(g_k)^2] \quad \text{and} \quad g_k = \left.\frac{\partial \mathcal{L}_{\widetilde{D}_k}(\boldsymbol{\theta})}{\partial \theta}\right|_{\boldsymbol{\theta}=\boldsymbol{\theta}_k} \tag{33}$$

in which $\beta_1$ and $\beta_2$ represent decay rates for the first and second-order moment estimates of the stochastic gradient with default values being $\beta_1 = 0.9$ and $\beta_2 = 0.999$ [43]; $m_k$ and $v_k$ are the first and second order moment estimates of stochastic gradient and are initialized by $m_0 = v_0 = 0$; $\epsilon_l$ = $10^{-8}$ is a small constant for numerical stability purposes; $\alpha_k$ is the learning rate at the current step and $\alpha$ is a hyperparameter for controlling the learning rate. Here, the normalized factor $\frac{1}{1-(\beta_1)^k}$ and $\frac{1}{1-(\beta_2)^k}$ are used to correct the moment estimates at the beginning of optimization and have little effect as $k$ becomes large.

### 3.3.4 Model parameter initialization

All the model parameters have to be initialized to start the model training process, i.e., parameter optimization, see Eq. (32). Unless transfer learning (Section 3.3.5) is used where the values of some model parameters are taken from other trained models, the initialization of the parameters is carried out as follows: (a) scaling and shifting parameters in all BN layers, $\boldsymbol{\gamma}$ and $\boldsymbol{\beta}$, are initialized to ones and zeros, respectively; (b) all the biases are initialized to zeros; (c) for all weights in CONV and FC layers, Xavier normal initialization method [44] is used, i.e., for any weight from a given neuron, it is randomly initialized from a Gaussian distribution $\mathcal{N}\left(0, \frac{2}{N_{\text{in}}}\right)$, where $N_{\text{in}}$ is the number of inputs for the this neuron. It should be noted that in a CONV layer, $N_{\text{in}}$ is equal to the filter size. With Xavier method, the variance of NN output is close to the variance of its input, which helps to prevent gradient vanishing/exploding issues at the start of training [44].

### 3.3.5 *Transfer learning*

a) From FC to SC layer

To construct a sparse neural network model described in Section 3.2, a sparsity analysis is carried out on the trained fully connected model in Section 3.1. Therefore, a fully connected model is first trained using the initialization strategy in Section 3.3.4 and following the training procedures described above. After a sparsity analysis on the weighting matrices with a pre-selected threshold, the remaining non-zero weights and biases are taken as the initialization of the corresponding parameters in the SC layers of the sparse network model, see Figure 1b. Starting from the basic neural network architecture with two CONV layers and transferred SC layers, the other model parameters in BN and CONV layers are initialized with the same strategy described in Section 3.3.4.

b) From shallow to deep CONV layer enriched SC networks

Starting with a basic architecture shown in Figure 3, more BN-CONV layers are added, as needed for a given problem. It is important to note that in the considered architecture, an SC layer has a significantly greater number of parameters as compared to a CONV layer and has a larger representational capacity. Therefore, it does not lead to a significant increase of trainable parameters by adding more BN-CONV layers. It should be remarked that although with Xavier initialization the gradient vanishing/exploding issues can be mitigated, the training performance can be still unstable, i.e., a deeper architecture may have larger training/testing error than a shallower architecture after the same number of training epochs. This is because even though a deeper network has more capacity, it is difficult to train a deeper network via a simple training strategy where all the parameters are simultaneously trained. To improve the training efficiency of deeper networks, the strategy of transfer learning [2] is utilized. The proposed transfer learning procedure for this study is detailed below.

In the proposed transfer learning strategy, the training starts with a basic NN architecture shown in Figure 3. The BN-CONV layers in the NN architecture are initialized by using the initialization procedure explained in Section 3.3.4, together with the transferred SC layer(s). After training, the values of all learned parameters are stored for future use. Suppose that more BN-CONV layer(s) are needed for higher representational capacity, then one CONV-BN layer is added between BN layer 1 and CONV 2, and the new architecture is shown in Figure 4. Using transfer learning, the new CONV layer 1-1 and BN layer 1-1 are trained in two steps. In the first step, all the trainable parameters from layers except the added CONV layer 1-1 and BN layer 1-1 are initialized by the ones stored in the previously trained architecture in Figure 3. The other trainable parameters are initialized in a standard way using Xavier initialization. In the first part of the training process, the parameters in all layers except the new layers are frozen and only the new layer parameters are trained. Thus, in this step, the previous knowledge is transferred to the new deeper architecture and this makes the training task easier by reducing the number of trainable parameters. In the second part of the training, all the layers are freed to continue training from where they stop in the first part of the training. This step is used to search the optimal solutions near the values obtained from the first step training such that better solutions are not missed. When more BN-CONV layers are used, they can be added one by one and trained by using the two-step strategy presented above.

In this manner, deeper architectures can effectively be trained to have smaller training errors. The efficacy of this training strategy is illustrated via numerical examples in the next section.

## 4    Numerical examples for SDOF systems

All the network training tasks are performed using TensorFlow on a superserver with dual twelve-core Intel Xeon© CPU @ 2.2GHz, 128GB RAM, and an Nvidia Geforce© GTX 1080 Ti GPU.

### 4.1   Dataspace

Five different input dataspaces are considered for the SDOF systems and are shown in Table 2. Among them, Case-1 to Case-4 are considered for linear SDOF problem (Section 2.1.1), while Case-5 is considered for nonlinear SDOF problem in Section 2.1.2. For MDOF systems, the input dataspaces for linear (Section 2.2.1) and nonlinear (Section 2.2.2) springs are considered in Case-6 and Case-7, respectively, in Table 2. Latin-hypercube sampling [45] is used to generate the input samples ($\boldsymbol{x}$) from the distributions of loading parameters $a_i$, $\omega_i$ and $\phi_i$ ($i = 1, 2, \cdots, N_p$), as it samples the entire dataspace more uniformly compared to random sampling. Output samples are then generated by solving the linear dynamical systems in Eq. (1) using analytical solutions in Appendix A.1 or nonlinear systems in Eq. (3) using the Newmark-$\beta$ method with a uniform time step size of 0.001 sec, which is tested to be able to give satisfactory accuracy [38]. The dataset that consists of input and output samples is divided into training ($N_{\text{train}}$), validation ($N_{\text{val}}$) and testing ($N_{\text{test}}$) datasets (Table 2), where *validation* dataset is used for checking overfitting and determining NN hyperparameters (Section 4.3.2) and *testing* dataset is used to test the performance of trained NN model for the unseen data, i.e., the prediction/generalization capacity of the model.

## 4.2 Hyperparameter settings

For NN models that employ only FC layers, i.e., nonlinear SDOF and MDOF dynamical systems, in Sections 4.5 and 5.2, as well as the SC layers in the sparse NN models, the number of neurons in each hidden layer is chosen to be the same as $n$ (the size of input/output). For the neural networks that exhibit sparsity pattern, the basic neural network architecture, as shown in Figure 3, comprises four computational layers: one CONV layer with $n_C^{[1]}$ filters, one CONV layer with 1 filter, two BN layers and transferred SC layers. More (BN-CONV) layers are added to this architecture to construct deeper networks, as needed. For all CONV layers, the filter size $f^{[l]}$ is fixed to be 2 and the stride $s^{[l]}$ is 1. The number of filters $n_C^{[l]}$ and the number of BN-CONV layers $n_l$ are two hyperparameters that are determined for each specific problem. In each BN layer, the hyperparameter $\epsilon_{BN}$ is fixed to be $10^{-8}$ for numerical stability.

For all the NN models training, the penalty factor in the loss function $\lambda = 10^{-4}$ is chosen for regularization using grid search from the set $\{10^{-6}, 10^{-5}, 10^{-4}, 10^{-3}, 10^{-2}, 10^{-1}\}$. For numerical stability purposes, the hyperparameter $\epsilon_D$ is fixed to be $10^{-8}$ in the training data normalization. For Adam optimizer, an optimal hyperparameter $\alpha = 10^{-3}$ is chosen based on a grid search from the set $\{10^{-5}, 10^{-4}, 10^{-3}, 10^{-2}, 10^{-1}\}$. All other hyperparameters in Adam optimizer are using recommended values listed in Section 3.3.3. When training the NN architecture, the batch size is fixed to 1024 and the number of epochs is fixed to 300 for all cases.

### 4.3 Linear SDOF system with a short loading duration

In this subsection, a short loading duration of $T = 5$ sec is considered, and this constitutes two cases – Case-1 and Case-2 – in Table 2. The two cases are used for investigating the influence of the number of harmonic frequencies in the loading on the model performance.

#### 4.3.1 Case-1: 5 sec loading with 5 harmonic terms

The number of harmonic terms in the loading is $N_p = 5$. The training dataset contains $2^{18} (= 262144)$ samples, while the validation dataset composes of 5000 samples. The testing dataset includes 5000 samples. In this case, a single dense layer is first trained to predict displacement or acceleration. The total number of trainable parameters is 10302. The histories of training and validation loss are shown in Figure 5. The corresponding training, validation, and testing mean squared errors are given in Table 3. Second, a sparsity analysis is carried out on the trained FC models, where a threshold value chosen as the 5% of the maximum absolute value of trained weight matrix $\boldsymbol{W}$ is used to filter out weights. The sparsity patterns of the FC layers after threshold cutoff are shown in Figure 6. Based on these sparsity patterns, the SC layer is constructed such that all the upper-triangular entries of the weight matrix in the SC layer are frozen to zeros and are not trained.

Next, the optimal number ($n_l$) of CONV layers and the optimal number of filters ($n_C^{[l]}$) in each CONV layer are investigated. In particular, the optimal number of filters ($n_C^{[l]}$) is selected from the set {1, 4, 8, 16, 32} by training two NN architectures with $n_l = 1$ and $n_l = 2$ for the displacement and acceleration responses, and the training and testing errors are given in Figure 7 and Figure 8. It can be seen that the errors are reducing as $n_C^{[l]}$ increases and the improvement slows down when $n_C^{[l]} \geq 16$. Therefore, $n_C^{[l]} = 16$ is chosen in all NN architectures in this study. To investigate the optimal $n_l$, more CONV layers are continuously added to the model. However, if random parameter initialization is used for deeper architectures, the performance of the trained model can be unstable, see Figure 9(a) where error increases when the number of CONV layers increases. To address this unstable training performance, transfer learning described in Section 3.3.5 is used, and accordingly, the training and testing errors for the same architectures are shown in Figure 9(b). With transfer learning, the overall error decreases when more CONV layers are added, and stable training performance is achieved. Thus, transfer learning will be considered in training NN architectures in all the following numerical cases. With transfer learning, the performance of displacement and acceleration models with different numbers of CONV layers is examined in more detail, see Figure 10 where $n_l = 6$ leads to an acceptable performance. Detailed information of the deep sparse NN architectures with $n_l = 6$ is summarized in Table 4. Compared to the baseline single dense layer network with 10302 trainable parameters, this sparse network contains only 9169 trainable parameters (11.0% reduction). However, the displacement and acceleration testing errors of the sparse architectures ($n_l = 6$) are $6.3472 \times 10^{-6}$ and $8.1845 \times 10^{-6}$, respectively, which are smaller than that of the FC layers ($3.3516 \times 10^{-5}$ and $3.3524 \times 10^{-5}$). Some predictions of sparse and baseline network models are compared against the analytical solutions in Figure 11, which again shows the competitive prediction accuracy of the sparse model as compared to the baseline dense network model.

### 4.3.2 Case-2: 5 sec loading with 25 harmonic terms

With other settings fixed to be the same as in Case-1, the number of harmonic terms is increased to $N_p = 25$ for Case-2 (Table 2). Following the same procedures described in Section 4.3.1, the histories of training and validation errors for the single FC layer model are shown in Figure 12. The sparsity pattern is obtained using the same threshold criterion as in Case-1 and a similar pattern is achieved, see Figure 13. Hence, the SC layer is again constructed with all the upper-triangular entries of the weight matrix in the SC layer frozen to zeros (non-trainable). The transferred SC layer is then combined with $n_l$ CONV layers ($n_c^{[l]} = 16$) and the training and testing errors for different $n_l$ values are shown in Figure 14(a) and (b) for displacement and acceleration, respectively. Again, $n_l = 6$ is chosen and results in the same NNs as in Table 4. The displacement and acceleration testing errors of sparse NNs ($n_l = 6$) are $6.4031 \times 10^{-6}$ and $6.2719 \times 10^{-6}$, respectively, and are smaller than that of the FC model, i.e., $3.1252 \times 10^{-5}$ and $3.3527 \times 10^{-5}$, respectively. Some predictions of sparse and baseline dense network models on testing samples are plotted in Figure 15, which shows similar prediction accuracy of the two models.

## 4.4  Linear SDOF system with a long loading duration

In this section, the performance of the proposed sparse NNs is investigated for a longer loading time duration, i.e., $T = 20$ sec. With the time step size $\Delta t = 0.1$ sec, this results in 201×1 input and output vectors, see Case-3 and Case-4 in Table 2.

### 4.4.1  Case-3: 20 sec loading with 5 harmonic terms

In Case-3 the applied load contains 5 harmonic terms. Following similar procedures in Section 4.3.2, a one-FC layer model is first trained with the results – training and validation loss – shown in Figure 16. Next, the sparsity patterns of the trained dense networks are obtained using the same criterion as Case-1 and are shown in Figure 17, which is different from Case-1 and Case-2 due to the changed time step size and loading duration. Based on that, an SC layer with a bandwidth of $w_b = 100$ in the lower-triangular part is constructed. With the transferred non-zero weights, the SC layer is then combined with CONV layers as before, and the results for different numbers ($n_l$) of added CONV layers are shown in Figure 18(a) and (b), respectively, for displacement and acceleration predictions. According to the results, $n_l = 9$ is chosen as they lead to desirable performance. The details of the final sparse network model are given in Table 5. Compared to the dense models, there is a 42.47% decrease in the number of trainable parameters in the sparse model. In the meantime, the testing errors of the sparse models are 0.0023 (for displacement) and 0.0025 (for acceleration), which are lower than that of the dense model, i.e., 0.0035 for displacement and 0.0047 for acceleration. Finally, some predictions on the testing samples of sparse and dense models are shown in Figure 19.

### 4.4.2  Case-4: 20 sec loading with 25 harmonic terms

Compared to Case-3, the number of harmonic terms, in this case, is increased to $N_p = 25$ in Case-4. With one FC layer, the dense models for the displacement and acceleration are first trained, see the results in Figure 20, and the trained layers are then used to obtain their sparsity patterns, see Figure 21. Due to the same sparsity pattern as compared to Case-3, the same structure of the SC layer is constructed but with different transferred learned weights from the FC layer. The training results of the transferred SC layer integrated with a different number ($n_l$) of CONV layers are given in Figure 22, where it can be seen that $n_l = 9$ leads to good accuracy. With $n_l = 9$, the resulted sparse NNs is the same as Case-3 of which the details are given in Table 5. This sparse NN leads to an over 42.47% decrease in the number of trainable parameters compared to the dense model. Meanwhile, the testing errors of the sparse model are 0.0024 (displacement) and 0.0025 (acceleration), which are smaller than that of the dense model, i.e. 0.0063 (displacement) and 0.0072 (acceleration). Some prediction samples from the testing dataset by the two models are presented in Figure 23.

## 4.5 Nonlinear SDOF system – Case-5

Considering long loading duration (20 sec) and 25 harmonic terms in the load, Case-5 is used to study the appropriate NN model for a nonlinear dynamical system. For the nonlinear case considering different values of cubic term coefficient $b$, a single FC layer is not enough for accurate prediction. Therefore, a hyperparameter study on the number of FC hidden layers (chosen from the set $\{1,2,...,7\}$) is carried out and the results are shown in Table 6, which shows the number of FC layers needed to reach the accuracy requirement that the mean relative errors of testing samples for predicting displacement and acceleration are smaller than 10%. Moreover, it is found that the number of FC layers needed to reach a good prediction accuracy increases as the system nonlinearity increases. Several prediction samples of displacement and acceleration by the dense models with an appropriate number of FC layers are shown in Figure 24, Figure 25, Figure 26, and Figure 27, which again confirm the good accuracy of the proposed dense models.

The sparsity patterns of the trained dense models are examined with the same criterion in Section 4.3.1 and *no sparsity patterns* are observed for both displacement and acceleration. As a result, no SC layers can be constructed. To leverage the lightweight property of CONV layers to reduce the size of the dense model, the first hidden FC layer is replaced with CONV layers. In the meantime, other layers in the trained dense model are transferred. It should be noted that the strategy of replacing multiple FC layers is not adopted here because it would result in much deeper CONV layers, and thus make the training more expensive. For illustration purposes, the replacement strategy is applied to the most challenging case where the cubic term coefficient $b = k$. The number ($n_l$) of added CONV layers is investigated by a parametric study with transfer learning. With $n_l = 5$, the mean relative errors of testing samples are 8.38% for displacement, and 8.65% for acceleration, which are close to that of the dense NN models, i.e., 8.63% for displacement and 8.76% for acceleration. However, the number of trainable parameters in the CONV enriched NNs models is 17.94% less than the original dense NNs models (see Table 7). The comparison of the predictions by the two types of models for several random testing samples is shown in Figure 28, where good prediction accuracy can be observed for both models.

## 5    Numerical examples for MDOF systems

A six-story multi-degree-of-freedom frame with Rayleigh damping is considered and the model details are given in Section 2.2. The two cases, which correspond to linear and cubic nonlinear springs, are considered in this study. In this problem, the input is the ground motion acceleration, and the output is the displacement or acceleration of the sixth degree-of-freedom with observation time step size being 0.1 sec. Detailed input dataset space is given in Table 2. The same hyperparameter settings given in Section 4.2 are used.

### 5.1   Case-6: Linear MDOF

The procedures for the CONV enriched sparse NNs training in Section 4.3 are also employed in this section. First, a series of FC layer-based dense models with the number of hidden layers from the set {1,2,3,4,5,6,7,8} are trained. With the harmonic loading frequencies drawn from the range $[0, 20\pi]$, however, increasing the hidden layers is not to be able to improve the accuracy, and all the trained NNs result in over 10% mean relative errors for both training and testing datasets. It is conjectured that due to the large dataspace, both the NN model complexity and the training datasets size need to be increased substantially to achieve successful training, and this puts a much higher demand on the computational hardware and requires much more computational effort to train.

To solve the MDOF problem with the available computational resources in a reasonable time window, the frequency range $[0, 20\pi]$ is divided into three parts, i.e., $[0,10\pi]$, $[10\pi, 15\pi]$ and $[15\pi, 20\pi]$. For each of the three dataspaces, a one-layer dense network is trained for the displacement or acceleration and its sparsity pattern is obtained using the same criterion in Section 4. The resulted sparsity patterns are shown in Figure 29, Figure 30, and Figure 31, respectively, for the three frequency ranges. The mean squared testing errors given in Table 8 and the mean relative errors given in Table 9 for the dense model on displacement and acceleration predictions show that with the data range division strategy the model accuracy becomes acceptable (less than 10% mean relative errors). Next, SC layers are constructed based on the sparsity patterns, which are similar to Case-2 and Case-4 in Sections 4.3.2 and 4.4.2. After that, the optimal number ($n_l$) of CONV layers to be added to SC layers is investigated and the training results with different $n_l$'s are plotted in Figure 32, Figure 33, and Figure 34. From the results, $n_l = 9$ is chosen for the sparse NNs model, which gives adequate accuracy. The mean squared errors and mean relative errors of sparse NN approximation for displacement and acceleration are given in Table 10 and Table 11. By comparing Table 8 with Table 10 and Table 9 with Table 11, it can be concluded that the proposed sparse NN architecture can be a good candidate for a parsimonious model with competitive accuracy. For different loading frequency ranges, predictions of displacement and acceleration on random loading samples from the testing dataset are compared against the analytical solutions in Figure 35, Figure 36, and Figure 37, where fairly good accuracy can be observed. It is noted that the combined three NNs trained for the three subranges $[0,10\pi]$, $[10\pi, 15\pi]$ and $[15\pi, 20\pi]$ cannot fully fulfill the original task where loading frequencies are drawn randomly from the single range $[0,2\pi]$. This is because the samples that contain frequencies across multiple subranges cannot be predicted using the three NNs.

## 5.2 Case-7: Nonlinear MDOF

Since nonlinear spring is considered in this case, a dense network with a single FC layer is unable to predict the responses accurately. Therefore, the hyperparameter study of dense networks in Section 4.5 is conducted in this section as well. The number of FC layers needed to reach an acceptable accuracy, i.e., mean relative testing error smaller than 10%, is shown in Table 12 and can be seen to increase with the nonlinearity in the system. Figure 38, Figure 39, Figure 40, and Figure 41 shows several prediction samples of displacement and acceleration for the dynamical system with different nonlinearities and these results confirm the good accuracy of the presented network models.

By examining the sparsity with the criterion in Section 4.3.1, it is found that there is no sparsity in the weight matrices in the dense network models, and therefore no sparse models can be constructed. Again, the same replacement strategy with lightweight CONV layers in Section 4.5 is adopted for the MDOF case with the largest nonlinearity, i.e., $b = k$. Transfer learning is used to investigate the number $(n_l)$ of CONV layers needed to reach a similar accuracy level as with the original dense layer. With $n_l = 5$ CONV layers, the mean relative errors of testing samples are 6.34% for displacement and 6.92% for acceleration, which are close to those from the original dense models, i.e., 6.86% for displacement and 7.00% for acceleration. Moreover, there is again a 17.94% decrease in the number of trainable parameters in the CONV enriched NN model as compared to the original dense model. Several predicted samples from the two NN models are shown in Figure 42, which confirms the competitive performance of the two NN architectures.

## 6 Conclusions

In this study, the use of deep neural network models as surrogates for linear and nonlinear structural dynamical systems is systematically examined. It is shown that for the considered linear SDOF and MDOF systems the fully connected neural network models exhibit sparsity patterns, i.e., some of the network weight parameters are much smaller than others. Moreover, the sparsity patterns can be different for different problem settings and may depend on the loading duration and the number of harmonic terms considered. This sparsity pattern can be utilized to prune a network architecture, which leads to lightweight and more memory-efficient network models. The performance of the sparse networks is further enhanced by adding parameter-efficient convolutional layers. The addition of the convolutional layers does not significantly increase the total number of trainable parameters due to their parameter sharing properties. As a result, the convolutional layer enhanced sparse network models are proposed for linear structural dynamics. The proposed sparse network models have competitive performance and are computationally efficient as compared to the fully connected NNs. Moreover, for training deep NN models, the transfer learning strategy proposed in this study is shown to have a stable training performance, i.e., loss function value continues to decrease as more convolutional layers are added.

For nonlinear dynamical response with cubic nonlinearities, it is shown that more fully connected network layers are needed as the magnitude of the cubic coefficient increases, i.e., a deeper network model is needed to learn a more complex system, as expected. Moreover, in this case, the sparsity patterns in the fully connected networks are not present. To take advantage of the lightweight property of CONV layers, several CONV layers are used to replace the first fully connected layer in the dense model and trained via transfer learning while other layers in the trained dense model are transferred. Compared to the fully connected NN model, the CONV enriched NN models have less trainable parameters while having similar testing performances. Thus, these models can be used as an alternative to fully connected NNs. Finally, in this study, only linear and nonlinear elastic systems are considered under harmonic excitations. Future work will explore the use of similar machine learning techniques for inelastic structural dynamical response prediction under more complex loading scenarios, e.g., seismic excitations and impact/blast loadings.

## ACKNOWLEDGMENTS

The presented work is supported in part by the USNational Science Foundation through grant CMMI-1762277. Any opinions, findings, conclusions, and recommendations expressed in this article are those of the authors and do not necessarily reflect the views of the sponsors.

# APPENDIX A

## ANALYTICAL SOLUTIONS OF LINEAR SDOF AND MDOF SYSTEMS

### A.1 Linear SDOF system

For linear SDOF systems in Eq. (1), the analytical solution of displacement can be expressed as

$$u(t) = u_g(t) + u_p(t)$$

(A 1)

where the general solution $u_g(t)$ and particular solution $u_p(t)$ are calculated as

$$u_g(t) = e^{-\varrho t}(A \cos \varsigma t + B \sin \varsigma t) \quad \text{with}$$

$$\varrho = \frac{c}{2m} \quad \text{and} \quad \varsigma = \frac{\sqrt{4mk - c^2}}{2m}$$

$$A = -\sum_{j=1}^{N_p} \frac{a_j}{k} \frac{-2\xi \frac{\omega_j}{\omega_n} \cos \phi_j + \left(1 - \left(\frac{\omega_j}{\omega_n}\right)^2\right) \sin \phi_j}{\left[1 - \left(\frac{\omega_j}{\omega_n}\right)^2\right]^2 + \left(2\xi \frac{\omega_j}{\omega_n}\right)^2}$$

(A 2)

$$B = \frac{1}{\varsigma}\left[\varrho A - \sum_{j=1}^{N_p} \frac{a_j \omega_j}{k} \frac{\left(1 - \left(\frac{\omega_j}{\omega_n}\right)^2\right) \cos \phi_j + 2\xi \frac{\omega_j}{\omega_n} \sin \phi_j}{\left[1 - \left(\frac{\omega_j}{\omega_n}\right)^2\right]^2 + \left(2\xi \frac{\omega_j}{\omega_n}\right)^2}\right]$$

and

$$u_p(t) = \sum_{j=1}^{N_p} \left(u_j^{p_1} + u_j^{p_2}\right) \quad \text{with}$$

$$u_j^{p_1} = C_j \sin \omega_j t + D_j \cos \omega_j t \quad \text{and} \quad u_j^{p_2} = E_j \sin \omega_j t + F_j \cos \omega_j t$$

(A 3)

$$C_j = \frac{a_j \cos \phi_j}{k} \frac{1 - \left(\frac{\omega_j}{\omega_n}\right)^2}{\left[1 - \left(\frac{\omega_j}{\omega_n}\right)^2\right]^2 + \left(2\xi \frac{\omega_j}{\omega_n}\right)^2}$$

$$D_j = \frac{a_j \cos \phi_j}{k} \frac{-2\xi \frac{\omega_j}{\omega_n}}{\left[1 - \left(\frac{\omega_j}{\omega_n}\right)^2\right]^2 + \left(2\xi \frac{\omega_j}{\omega_n}\right)^2}$$

$$E_j = \frac{a_j \sin \phi_j}{k} \frac{2\xi \frac{\omega_j}{\omega_n}}{\left[1 - \left(\frac{\omega_j}{\omega_n}\right)^2\right]^2 + \left(2\xi \frac{\omega_j}{\omega_n}\right)^2}$$

$$F_j = \frac{a_j \sin \phi_j}{k} \frac{1 - \left(\frac{\omega_j}{\omega_n}\right)^2}{\left[1 - \left(\frac{\omega_j}{\omega_n}\right)^2\right]^2 + \left(2\xi \frac{\omega_j}{\omega_n}\right)^2}$$

The acceleration can be computed by taking the second order derivatives of $u(t)$ in Eq. (A 1), i.e.

$$\ddot{u}(t) = \ddot{u}_g(t) + \ddot{u}_p(t) \tag{A 4}$$

where

$$\ddot{u}_g(t) = \varrho^2 e^{-\varrho t}(A \cos \varsigma t + B \sin \varsigma t) - 2\varrho e^{-\varrho t}(-A\varsigma \sin \varsigma t + B\varsigma \cos \varsigma t)$$
$$+ e^{-\varrho t}(-A\varsigma^2 \cos \varsigma t - B\varsigma^2 \sin \varsigma t) \tag{A 5}$$

and

$$\ddot{u}_p(t) = \sum_{j=1}^{N_p} \left(\ddot{u}_j^{p_1} + \ddot{u}_j^{p_2}\right) \quad \text{with}$$

$$\ddot{u}_j^{p_1} = -C_j \omega_j^2 \sin(\omega_j t) - D_j \omega_j^2 \cos(\omega_j t) \tag{A 6}$$

$$\ddot{u}_j^{p_2} = -E_j \omega_j^2 \sin(\omega_j t) - F_j \omega_j^2 \cos(\omega_j t)$$

A.2 Linear MDOF system

For linear MDOF system in Eq. (4), the analytical solution of displacement given by modal decomposition reads

$$u(t) = \widetilde{\phi}q(t) \quad \text{with} \quad \widetilde{\phi} = [\widetilde{\phi}_1, \widetilde{\phi}_2, \widetilde{\phi}_3, \widetilde{\phi}_4, \widetilde{\phi}_5, \widetilde{\phi}_6] \quad \text{and}$$

$$q(t) = \begin{bmatrix} q_1(t) \\ q_2(t) \\ q_3(t) \\ q_4(t) \\ q_5(t) \\ q_6(t) \end{bmatrix} \quad \text{and} \quad \widetilde{\phi}_i \overset{\text{def}}{=} \frac{1}{\sqrt{\phi_i^T M \phi_i}} \phi_i, i = 1,2,\dots 6 \tag{A 7}$$

where $(\omega_{ni}^2, \phi_i)$ are eigenvalue-eigenvector pairs of the matrix $M^{-1}K$ and

$$q_i(t) = q_i^g(t) + q_i^p(t), i = 1,2,\dots 6 \tag{A 8}$$

with the general solution

$$q_i^g(t) = e^{-\varrho_i t}(G_i \cos \varsigma_i t + H_i \sin \varsigma_i t) \quad \text{with}$$

$$G_i = -\sum_{j=1}^{N_p} \frac{\gamma_i a_j}{\omega_{ni}^2} \left( \frac{-2\zeta_i \frac{\omega_j}{\omega_{ni}} \cos \phi_j + \left(1 - \left(\frac{\omega_j}{\omega_{ni}}\right)^2\right) \sin \phi_j}{\left[1 - \left(\frac{\omega_j}{\omega_{ni}}\right)^2\right]^2 + \left(2\zeta_i \frac{\omega_j}{\omega_{ni}}\right)^2} \right)$$

$$H_i = \frac{1}{\varsigma_i} \left[ \varrho_i A_i - \sum_{j=1}^{N_p} \frac{\gamma_i a_j \omega_j}{\omega_{ni}^2} \left( \frac{\left(1 - \left(\frac{\omega_j}{\omega_{ni}}\right)^2\right) \cos \phi_j + 2\zeta_i \frac{\omega_j}{\omega_{ni}} \sin \phi_j}{\left[1 - \left(\frac{\omega_j}{\omega_{ni}}\right)^2\right]^2 + \left(2\zeta_i \frac{\omega_j}{\omega_{ni}}\right)^2} \right) \right] \tag{A 9}$$

$$\varrho_i = \frac{\alpha_0 + \alpha_1 \omega_{ni}^2}{2} \quad \text{and} \quad \varsigma_i = \frac{\sqrt{4\omega_{ni}^2 - (\alpha_0 + \alpha_1 \omega_{ni}^2)^2}}{2}$$

$$\gamma_i = -\widetilde{\phi}_i^T \text{diag}(M) \quad \text{and} \quad \zeta_i = \frac{\alpha_0 + \alpha_1 \omega_{ni}^2}{2\omega_{ni}}$$

and particular solution

$$q_i^p(t) = \sum_{j=1}^{N_p} \left( q_i^{p_{1j}} + q_i^{p_{2j}} \right) \quad \text{with} \tag{A 10}$$

$$q_i^{p_{1j}}(t) = I_i \sin(\omega_j t) + J_i \cos(\omega_j t)$$

$$q_i^{p_{2j}}(t) = L_i \sin(\omega_j t) + M_i \cos(\omega_j t)$$

$$I_i = \frac{\gamma_i a_j \cos\phi_j}{\omega_{ni}^2} \frac{1 - \left(\frac{\omega_j}{\omega_{ni}}\right)^2}{\left[1 - \left(\frac{\omega_j}{\omega_{ni}}\right)^2\right]^2 + \left(2\zeta_i \frac{\omega_j}{\omega_{ni}}\right)^2}$$

$$J_i = \frac{\gamma_i a_j \cos\phi_j}{\omega_{ni}^2} \frac{-2\zeta_i \frac{\omega_j}{\omega_{ni}}}{\left[1 - \left(\frac{\omega_j}{\omega_{ni}}\right)^2\right]^2 + \left(2\zeta_i \frac{\omega_j}{\omega_{ni}}\right)^2}$$

$$L_i = \frac{\gamma_i a_j \sin\phi_j}{\omega_{ni}^2} \frac{2\zeta_i \frac{\omega_j}{\omega_{ni}}}{\left[1 - \left(\frac{\omega_j}{\omega_{ni}}\right)^2\right]^2 + \left(2\zeta_i \frac{\omega_j}{\omega_{ni}}\right)^2}$$

$$M_i = \frac{\gamma_i a_j \sin\phi_j}{\omega_{ni}^2} \frac{1 - \left(\frac{\omega_j}{\omega_{ni}}\right)^2}{\left[1 - \left(\frac{\omega_j}{\omega_{ni}}\right)^2\right]^2 + \left(2\zeta_i \frac{\omega_j}{\omega_{ni}}\right)^2}$$

Taking the second order derivative of $\boldsymbol{u}(t)$ in Eq. (A 7), the analytical solution of acceleration reads

$$\ddot{\boldsymbol{u}}(t) = \widetilde{\boldsymbol{\phi}} \ddot{\boldsymbol{q}}(t) \text{ with}$$

$$\ddot{\boldsymbol{q}}(t) = \begin{bmatrix} \ddot{q}_1(t) \\ \ddot{q}_2(t) \\ \ddot{q}_3(t) \\ \ddot{q}_4(t) \\ \ddot{q}_5(t) \\ \ddot{q}_6(t) \end{bmatrix} \tag{A 11}$$

where

$$\ddot{q}_i(t) = \ddot{q}_i^g(t) + \ddot{q}_i^p(t), \qquad i = 1,2,\dots 6, \qquad \text{with} \tag{A 12}$$

$$\ddot{q}_i^g(t) = \varrho_i^2 e^{-\varrho_i t}(G_i \cos\varsigma_i t + H_i \sin\varsigma_i t) - 2\varrho e^{-\varrho t}(-G_i\varsigma_i \sin\varsigma_i t + H_i\varsigma_i \cos\varsigma_i t)$$

$$+ e^{-\varrho_i t}(-G_i\varsigma_i^2 \cos\varsigma_i t - H_i\varsigma_i^2 \sin\varsigma_i t)$$

$$\ddot{q}_i^p(t) = \sum_{j=1}^{N_p}(\ddot{q}_i^{p_1 j} + \ddot{q}_i^{p_2 j})$$

$$\ddot{q}_i^{p_1 j} = -I_i\omega_j^2 \sin(\omega_j t) - J_i\omega_j^2 \cos(\omega_j t)$$

$$\ddot{q}_i^{p_2 j} = -L_i\omega_j^2 \sin(\omega_j t) - M_i\omega_j^2 \cos(\omega_j t)$$

Table 1. Summary of model parameters in MDOF system

| $m_1$ (kg) | $m_2$ (kg) | $m_3$ (kg) | $E$ (N/m$^2$) | $I$ (m$^4$) | $L$ (m) | $\xi$ | $\omega_{n1}$ (rad/s) | $\omega_{n2}$ (rad/s) |
|---|---|---|---|---|---|---|---|---|
| 14000 | 12000 | 10000 | $2 \times 10^{11}$ | $4.2 \times 10^{-4}$ | 3.5 | 0.02 | 15.73 | 44.16 |

Table 2. Summary of input dataspaces for different cases

| Case # | Input dataspace parameters | | |
|---|---|---|---|
| Case-1 | $\mathcal{D}_{\text{lsdof}} = \{a_i \sim \mathcal{U}(0,10),$ <br> $\varphi_i \sim \mathcal{U}(0,2\pi),$ <br> $\omega_i \sim \mathcal{U}(0,6\pi),\ i = 1,2,\cdots,N_p\},\ N_p = 5$ | $N_{\text{train}} = 2^{18},$ <br> $N_{\text{test}} = 5000$ <br> $N_{\text{val}} = 5000$ | $T = 5$ sec, <br> $\Delta t = 0.05$ sec <br> $n = 101$ |
| Case-2 | $\mathcal{D}_{\text{lsdof}} = \{a_i \sim \mathcal{U}(0,10),$ <br> $\varphi_i \sim \mathcal{U}(0,2\pi),$ <br> $\omega_i \sim \mathcal{U}(0,6\pi),\ i = 1,2,\cdots,N_p\},\ N_p = 25$ | $N_{\text{train}} = 2^{18},$ <br> $N_{\text{test}} = 5000$ <br> $N_{\text{val}} = 5000$ | $T = 5$ sec, <br> $\Delta t = 0.05$ sec <br> $n = 101$ |
| Case-3 | $\mathcal{D}_{\text{lsdof}} = \{a_i \sim \mathcal{U}(0,10),$ <br> $\varphi_i \sim \mathcal{U}(0,2\pi),$ <br> $\omega_i \sim \mathcal{U}(0,6\pi),\ i = 1,2,\cdots,N_p\},\ N_p = 5$ | $N_{\text{train}} = 2^{18},$ <br> $N_{\text{test}} = 5000$ <br> $N_{\text{val}} = 5000$ | $T = 20$ sec, <br> $\Delta t = 0.1$ sec <br> $n = 201$ |
| Case-4 | $\mathcal{D}_{\text{lsdof}} = \{a_i \sim \mathcal{U}(0,10),$ <br> $\varphi_i \sim \mathcal{U}(0,2\pi),$ <br> $\omega_i \sim \mathcal{U}(0,6\pi),\ i = 1,2,\cdots,N_p\},\ N_p = 25$ | $N_{\text{train}} = 2^{18},$ <br> $N_{\text{test}} = 5000$ <br> $N_{\text{val}} = 5000$ | $T = 20$ sec, <br> $\Delta t = 0.1$ sec <br> $n = 201$ |
| Case-5 | $\mathcal{D}_{\text{nlsdof}} = \{a_i \sim \mathcal{U}(0,20),$ <br> $\varphi_i \sim \mathcal{U}(0,2\pi),$ <br> $\omega_i \sim \mathcal{U}(0,6\pi),\ i = 1,2,\cdots,N_p\},\ N_p = 25$ | $N_{\text{train}} = 2^{20},$ <br> $N_{\text{test}} = 5000$ <br> $N_{\text{val}} = 5000$ | $T = 20$ sec, <br> $\Delta t = 0.1$ sec <br> $n = 201$ |
| Case-6 | $\mathcal{D}_{\text{mdof}} = \{a_i \sim \mathcal{U}(0,4),$ <br> $\varphi_i \sim \mathcal{U}(0,2\pi),$ <br> $\omega_i \sim \mathcal{U}(0,20\pi),\ i = 1,2,\cdots,N_p\},\ N_p = 25$ | $N_{\text{train}} = 2^{20},$ <br> $N_{\text{test}} = 5000$ <br> $N_{\text{val}} = 5000$ | $T = 20$ sec, <br> $\Delta t = 0.1$ sec <br> $n = 201$ |
| Case-7 | $\mathcal{D}_{\text{mdof}} = \{a_i \sim \mathcal{U}(0,4),$ <br> $\varphi_i \sim \mathcal{U}(0,2\pi),$ <br> $\omega_i \sim \mathcal{U}(0,10\pi),\ i = 1,2,\cdots,N_p\},\ N_p = 25$ | $N_{\text{train}} = 2^{20},$ <br> $N_{\text{test}} = 5000$ <br> $N_{\text{val}} = 5000$ | $T = 20$ sec, <br> $\Delta t = 0.1$ sec <br> $n = 201$ |

Table 3. Training, validation and testing mean squared errors of the trained dense networks for linear SDOF system (Case-1)

| | Displacement | Acceleration |
|---|---|---|
| Training error | $3.0759 \times 10^{-5}$ | $3.1964 \times 10^{-5}$ |
| Validation error | $3.1658 \times 10^{-5}$ | $3.2958 \times 10^{-5}$ |
| Testing error | $3.3516 \times 10^{-5}$ | $3.3524 \times 10^{-5}$ |

Table 4. Summary of parameters in the sparse network architecture ($n_l = 6$) for Case-1 and Case-2

| | Layer type | # of trainable variables | Activation | | Layer type | # of trainable variables | Activation |
|---|---|---|---|---|---|---|---|
| Layer 1 | Convolution | 48 | ReLU | Layer 8 | BN | 502 | None |
| Layer 2 | BN | 404 | None | Layer 9 | Convolution | 528 | ReLU |
| Layer 3 | Convolution | 528 | ReLU | Layer 10 | BN | 502 | None |
| Layer 4 | BN | 404 | None | Layer 11 | Convolution | 528 | ReLU |
| Layer 5 | Convolution | 528 | ReLU | Layer 12 | BN | 502 | None |
| Layer 6 | BN | 404 | None | Layer 13 | Convolution | 17 | None |
| Layer 7 | Convolution | 528 | ReLU | Layer 14 | SC | 5252 | None |
| Total # of trainable parameters: 9169 | | | | | | | |
| Reduced percentage of trainable parameters compared to dense model: 11.00% | | | | | | | |

Table 5. Summary of parameters in the sparse network architecture ($n_l = 9$) for Case-3, Case-4, and Case-6

| | Layer type | # of trainable variables | Activation | | Layer type | # of trainable variables | Activation |
|---|---|---|---|---|---|---|---|
| Layer 1 | Convolution | 48 | ReLU | Layer 11 | Convolution | 528 | ReLU |
| Layer 2 | BN | 804 | None | Layer 12 | BN | 804 | None |
| Layer 3 | Convolution | 528 | ReLU | Layer 13 | Convolution | 528 | ReLU |
| Layer 4 | BN | 804 | None | Layer 14 | BN | 804 | None |
| Layer 5 | Convolution | 528 | ReLU | Layer 15 | Convolution | 528 | ReLU |
| Layer 6 | BN | 804 | None | Layer 16 | BN | 804 | None |
| Layer 7 | Convolution | 528 | ReLU | Layer 17 | Convolution | 528 | ReLU |
| Layer 8 | BN | 804 | None | Layer 18 | BN | 804 | None |
| Layer 9 | Convolution | 528 | ReLU | Layer 19 | Convolution | 17 | None |
| Layer 10 | BN | 804 | None | Layer 20 | SC | 15452 | None |
| Total # of trainable parameters: 23359 | | | | | | | |
| Reduced percentage of trainable parameters compared to dense model: 42.47% | | | | | | | |

Table 6. Number of needed FC layers in the dense networks for nonlinear SDOF system (Case-5) to reach acceptable accuracy (mean relative error of testing samples smaller than 10%) and the corresponding mean relative errors

| Cubic term coefficient $b$ | Number of needed FC layers | Displacement | Acceleration |
|---|---|---|---|
| $0.25k$ | 2 | 7.56% | 7.97% |
| $0.5k$ | 2 | 7.74% | 8.88% |
| $0.75k$ | 4 | 7.44% | 7.92% |
| $k$ | 5 | 8.63% | 8.76% |

Table 7. Summary of parameters in the CONV layers enriched network architecture ($n_l = 5$) for Case-5 and Case-7

| | Layer type | # of trainable variables | Activation | | Layer type | # of trainable variables | Activation |
|---|---|---|---|---|---|---|---|
| Layer 1 | Convolution | 48 | ReLU | Layer 9 | Convolution | 528 | ReLU |
| Layer 2 | BN | 804 | None | Layer 10 | BN | 804 | None |
| Layer 3 | Convolution | 528 | ReLU | Layer 11 | Convolution | 17 | None |
| Layer 4 | BN | 804 | None | Layer 12 | FC | 40602 | ReLU |
| Layer 5 | Convolution | 528 | ReLU | Layer 13 | FC | 40602 | ReLU |
| Layer 6 | BN | 804 | None | Layer 14 | FC | 40602 | ReLU |
| Layer 7 | Convolution | 528 | ReLU | Layer 15 | FC | 40602 | None |
| Layer 8 | BN | 804 | None | | | | |

| Total # of trainable parameters: 166595 |
|---|
| Percentage reduction in trainable parameters compared to dense model: 17.94% |

Table 8. Mean squared testing errors of the trained dense networks for linear MDOF system (Case-6) with different loading frequency ranges

| Loading frequency range | Displacement | Acceleration |
|---|---|---|
| $[0,10\pi]$ | 0.0049 | 0.0051 |
| $[10\pi, 15\pi]$ | 0.0020 | 0.0094 |
| $[15\pi, 20\pi]$ | 0.0084 | 0.0183 |

Table 9. Mean relative errors over testing dataset of the trained dense networks for linear MDOF system (Case-6) with different loading frequency ranges

| Loading frequency range | Displacement | Acceleration |
|---|---|---|
| $[0,10\pi]$ | 1.27% | 1.30% |
| $[10\pi, 15\pi]$ | 0.98% | 6.68% |
| $[15\pi, 20\pi]$ | 6.21% | 8.96% |

Table 10. Mean squared testing errors of the trained sparse network ($n_l = 9$) for MDOF system (Case-6) with different loading frequency ranges

| Loading frequency range | Displacement | Acceleration |
|---|---|---|
| $[0,10\pi]$ | 0.0051 | 0.0064 |
| $[10\pi, 15\pi]$ | 0.0022 | 0.0098 |
| $[15\pi, 20\pi]$ | 0.0069 | 0.0159 |

Table 11. Mean relative errors over testing dataset of the trained sparse network ($n_l = 9$) for MDOF system (Case-6) with different loading frequency ranges

| Loading frequency range | Displacement | Acceleration |
|---|---|---|
| $[0, 10\pi]$ | 3.45% | 3.53% |
| $[10\pi, 15\pi]$ | 1.33% | 6.84% |
| $[15\pi, 20\pi]$ | 5.49% | 7.68% |

Table 12. Number of needed FC layers of the trained dense networks for nonlinear MDOF system (Case-7) to reach the acceptable accuracy (mean relative error of testing samples smaller than 10%) and the corresponding mean relative errors

| Cubic term coefficient $b$ | Number of needed FC layers | Displacement | Acceleration |
|---|---|---|---|
| $0.25k$ | 2 | 6.11% | 6.31% |
| $0.5k$ | 2 | 7.04% | 7.72% |
| $0.75k$ | 4 | 6.41% | 6.88% |
| $k$ | 5 | 6.86% | 7.00% |

FC layer

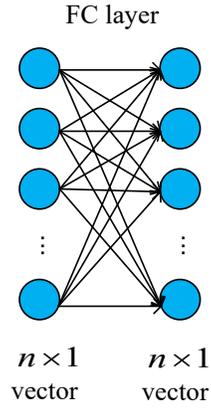

$n \times 1$
vector

$n \times 1$
vector

(a) Dense network

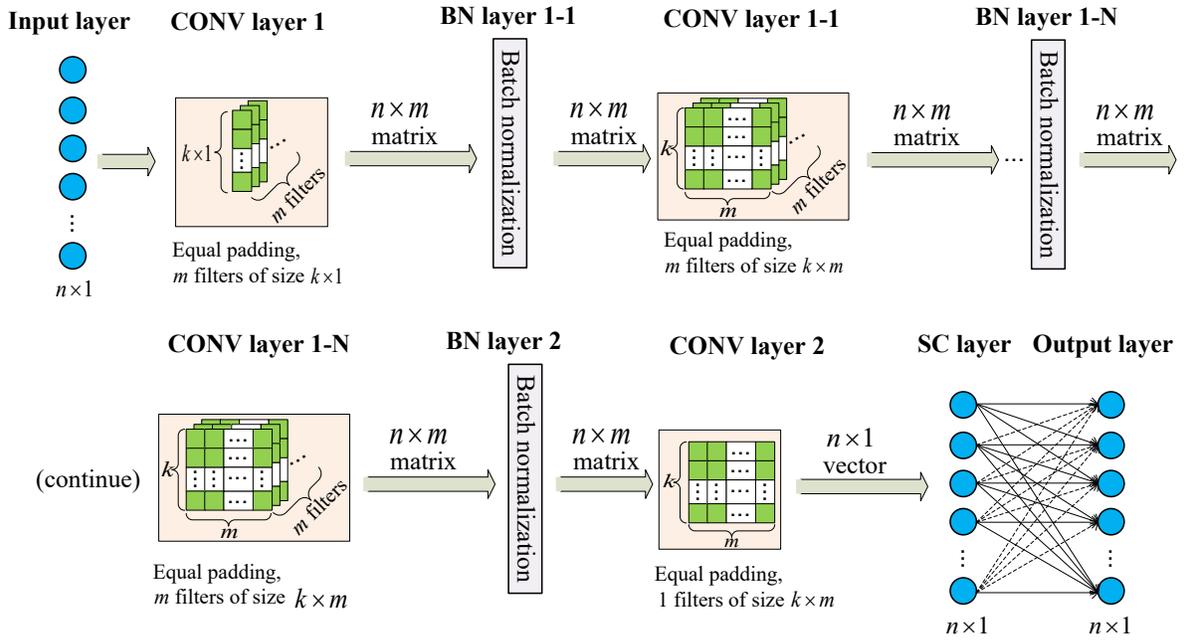

(b) Sparse network architecture

Figure 1. Neural network architectures.

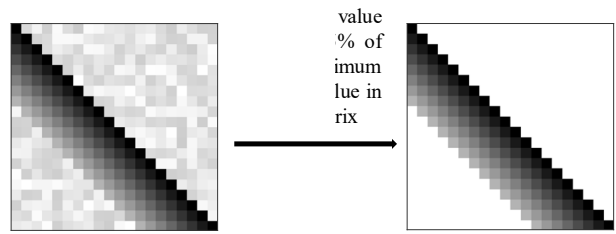

Figure 2. Illustration of sparsity pattern.

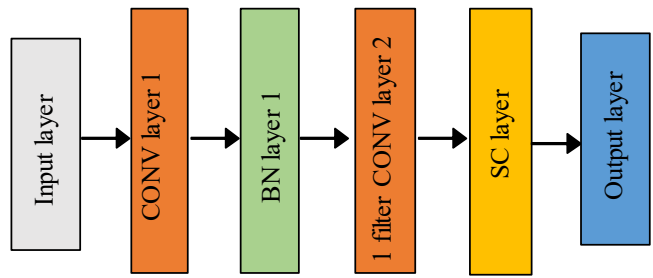

Figure 3. Basic architecture of the CONV enriched sparse NNs.

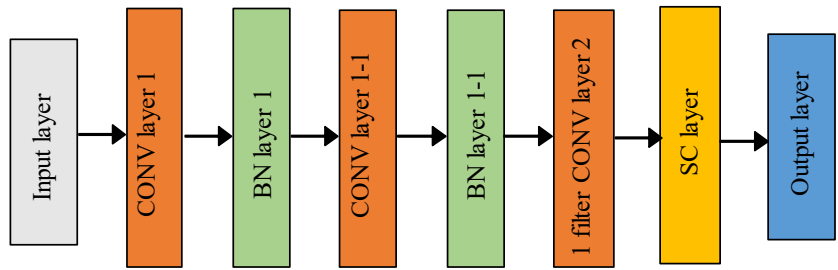

Figure 4. Deep CONV enriched sparse NNs with added BN-CONV layers.

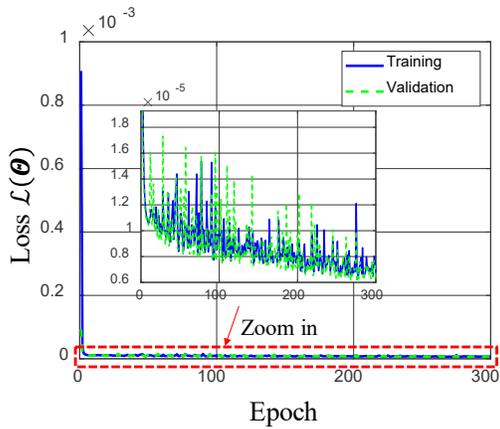 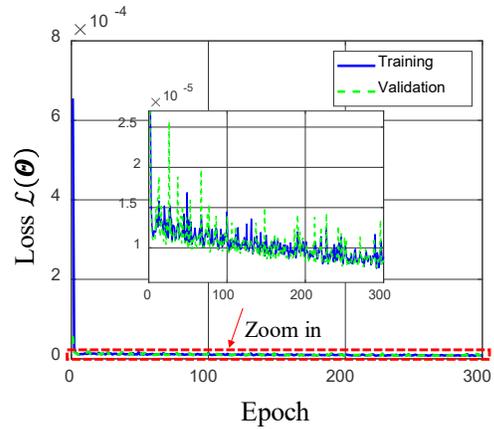

(a) Displacement            (b) Acceleration

Figure 5. Training and validation loss histories during training process of the dense FC model in Case-1: (a) Displacement; (b) Acceleration.

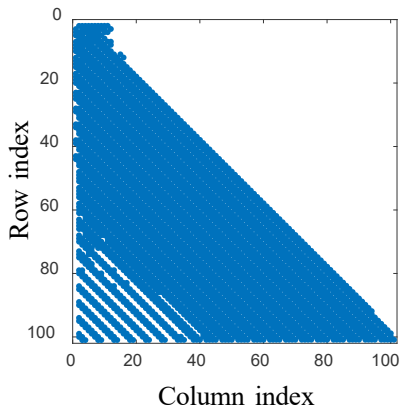 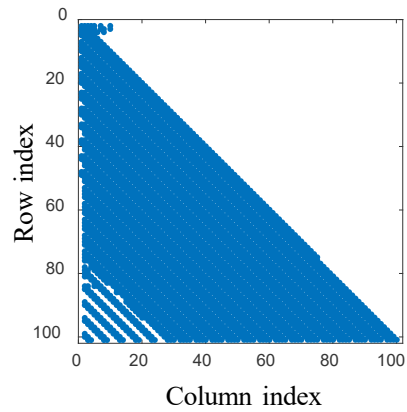

(a) Displacement (#non-zero weights = 3979)     (b) Acceleration (#non-zero weights = 3985)

Figure 6. Sparsity patterns of the weight matrix of (single) FC layer in the dense models in Case-1: (a) Displacement; (b) Acceleration.

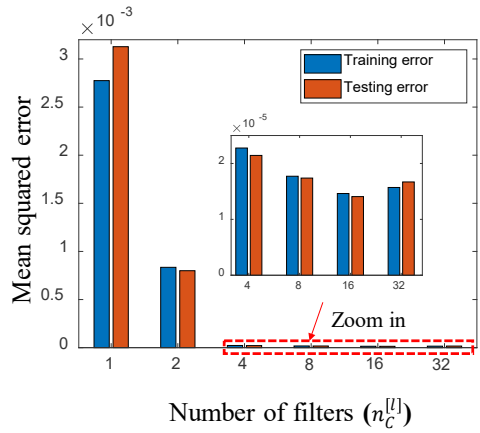
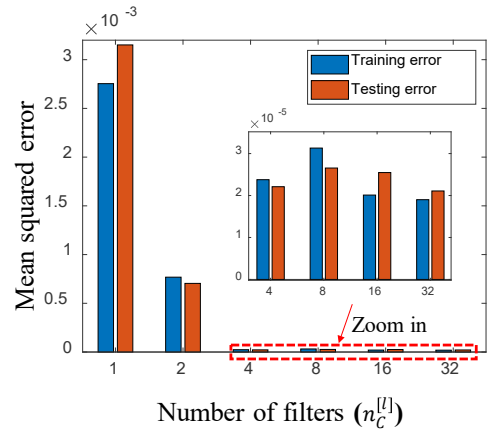

(a) $\boldsymbol{n_l} = 1$          (b) $\boldsymbol{n_l} = 2$

Figure 7. Parametric studies of the number of filters ($n_c^{[l]}$) in CONV layers in Case-1 for displacement.

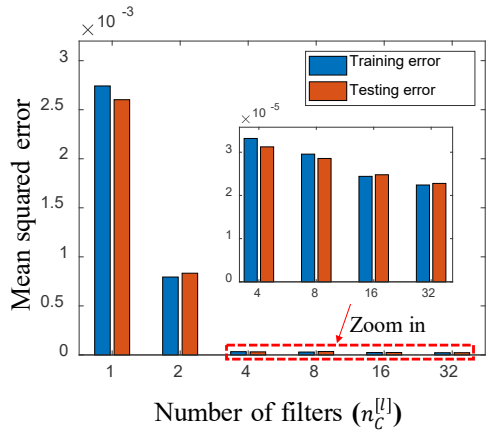
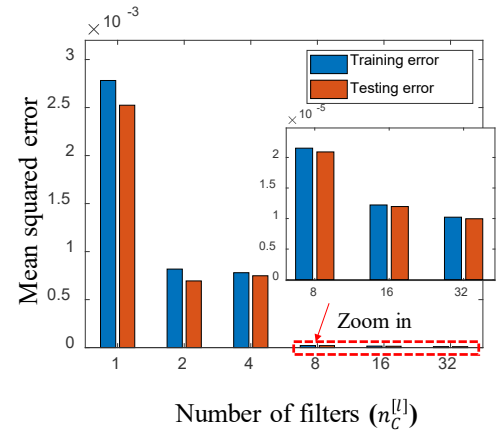

(a) $\boldsymbol{n_l} = 1$          (b) $\boldsymbol{n_l} = 2$

Figure 8. Parametric studies of the number of filters ($n_c^{[l]}$) in CONV layers in Case-1 for acceleration.

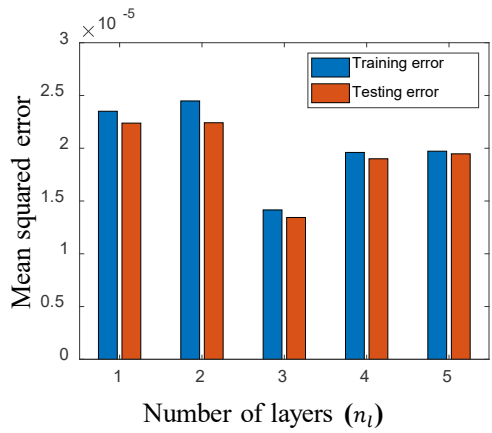

(a) Random initialization

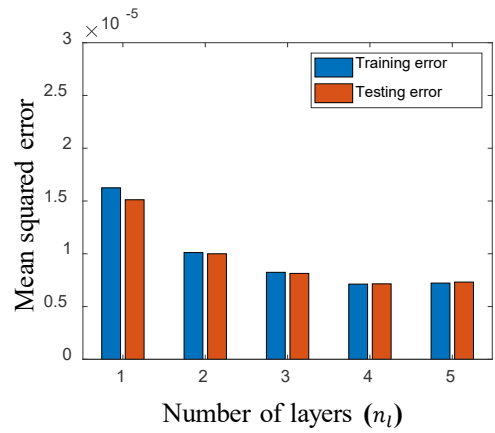

(b) Initialization by transfer learning

Figure 9. Comparison of different training strategies on multiple networks of displacement model (Case-1).

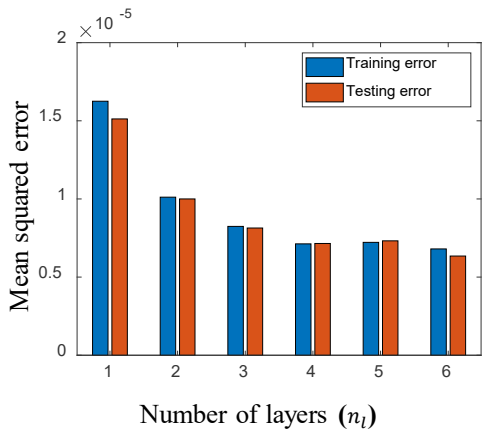

(a) Displacement

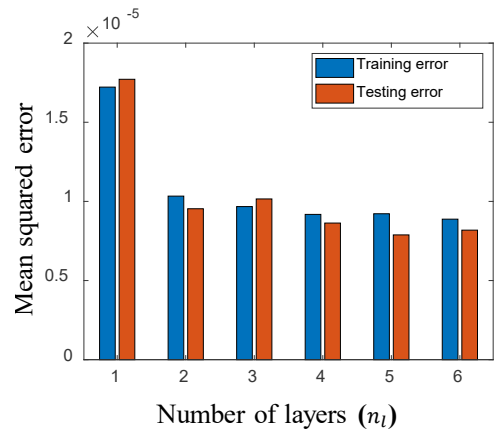

(b) Acceleration

Figure 10. Parametric studies on the optimal number ($n_l$) of CONV layers in the sparse networks in Case-1: (a) Displacement; (b) Accelerations.

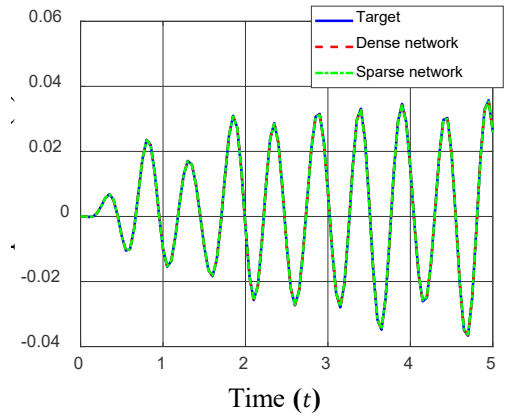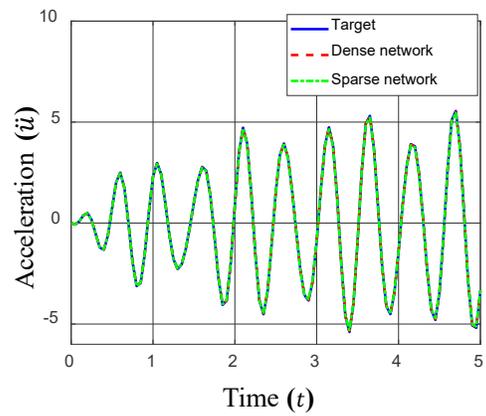

(a) Sample 1

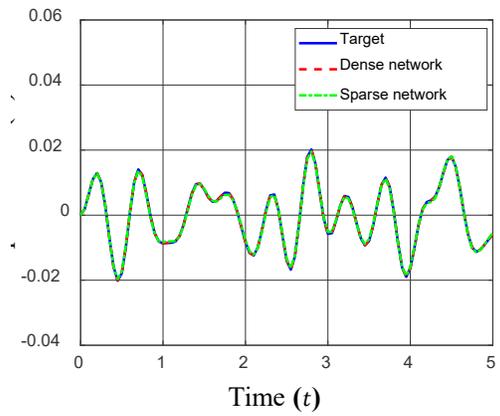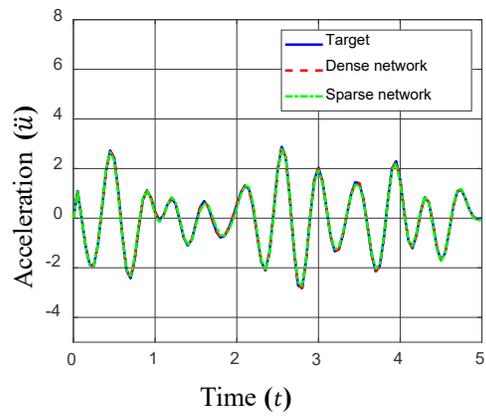

(b) Sample 2

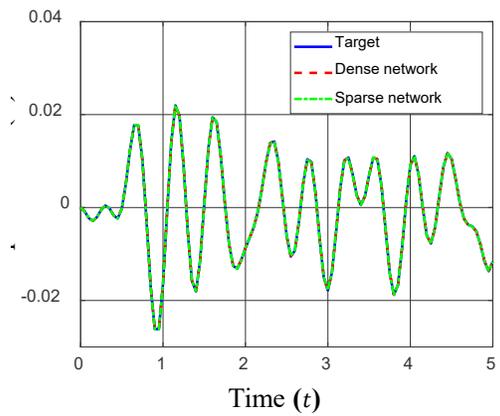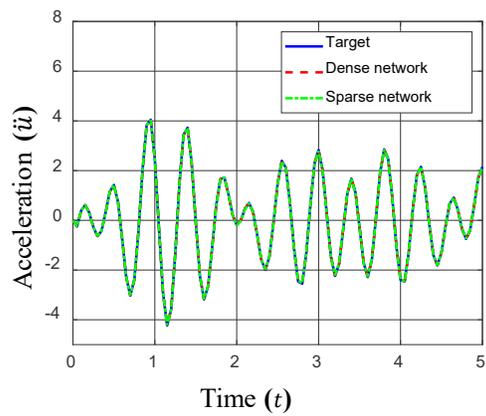

(c) Sample 3

Figure 11. Performance comparison of the dense and sparse network models on three random samples from testing dataset in Case-1: Left – displacement; Right – acceleration.

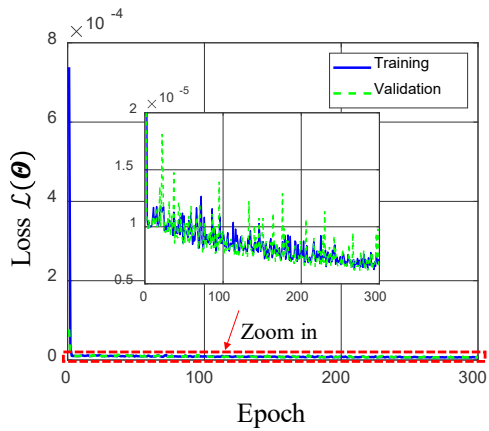 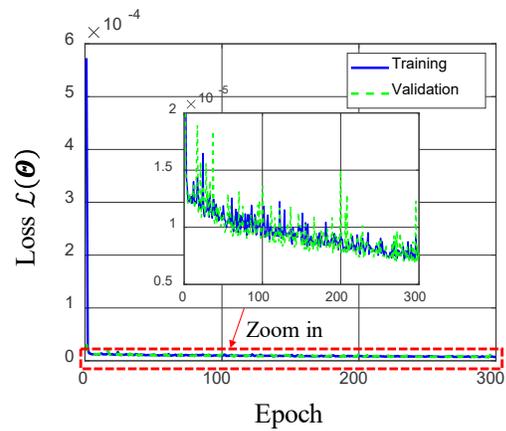

(a) Displacement            (b) Acceleration

Figure 12. Training and validation loss histories during training process of
the FC dense model in Case-2: (a) Displacement; (b) Acceleration.

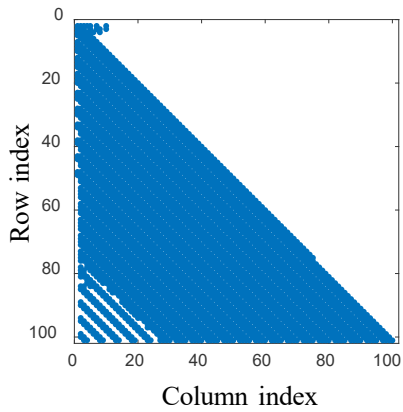 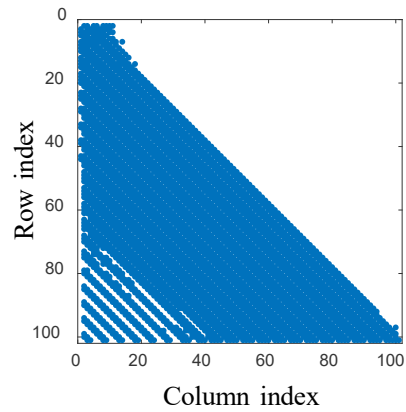

(a) Displacement (#non-zero weights = 3976)     (b) Acceleration (#non-zero weights = 3985)

Figure 13. Sparsity patterns of the weight matrix of (single) FC layer in
the dense models in Case-2: (a) Displacement; (b) Acceleration.

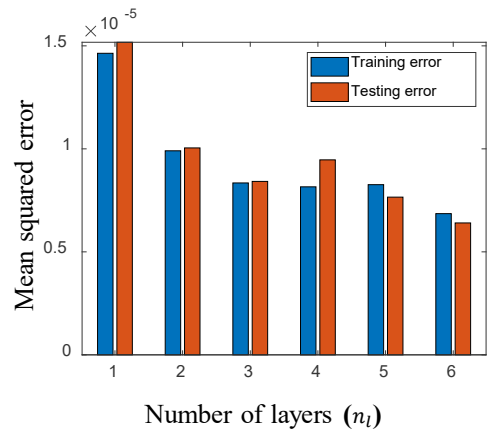
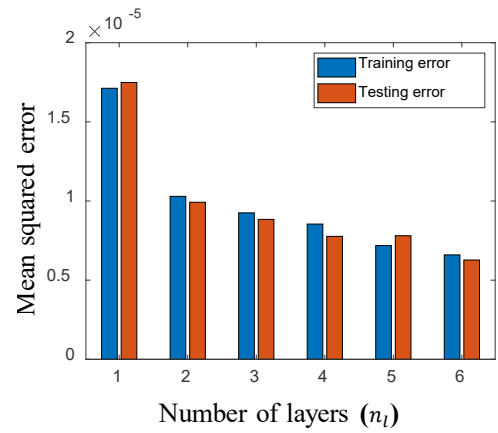

(a) Displacement            (b) Acceleration

Figure 14. Parametric studies on the optimal number ($n_l$) of CONV layers in the sparse networks in Case-2: (a) Displacement; (b) Acceleration.

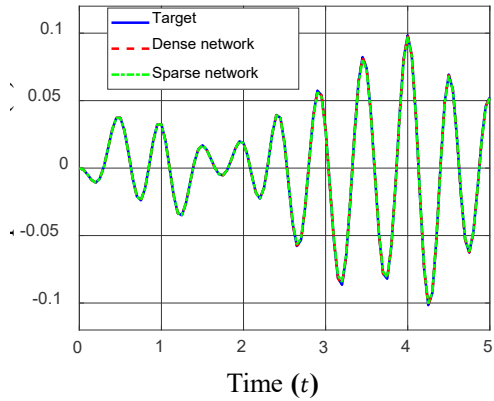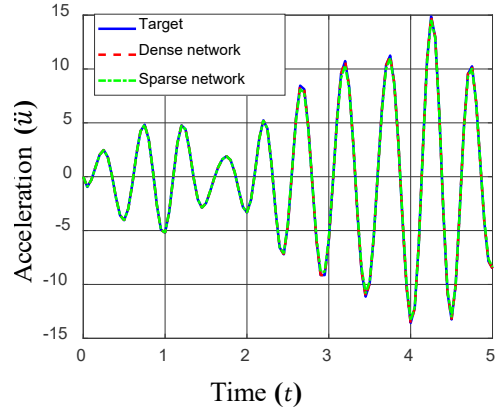

(a) Sample 1

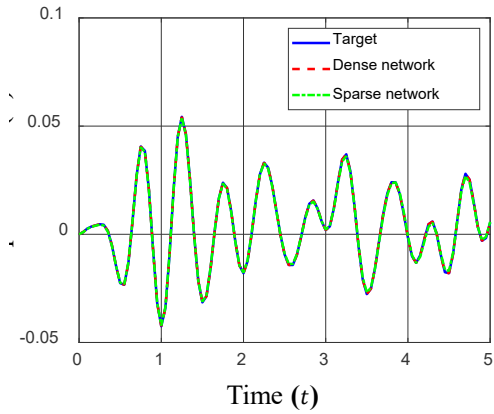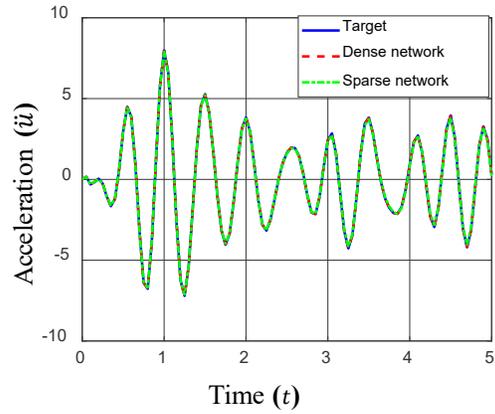

(b) Sample 2

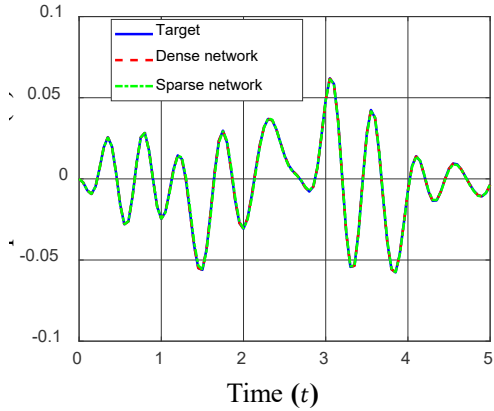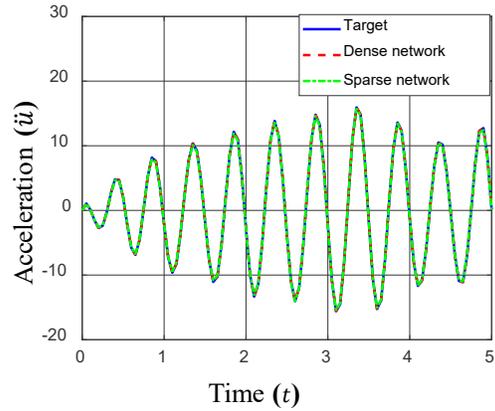

(c) Sample 3

Figure 15. Performance comparison of the dense and sparse network models on three random samples from testing dataset in Case-2: Left – displacement; Right – acceleration.

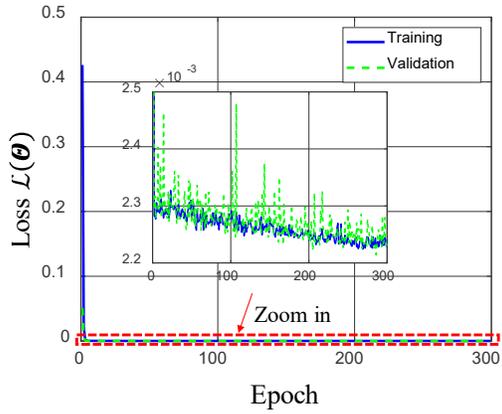 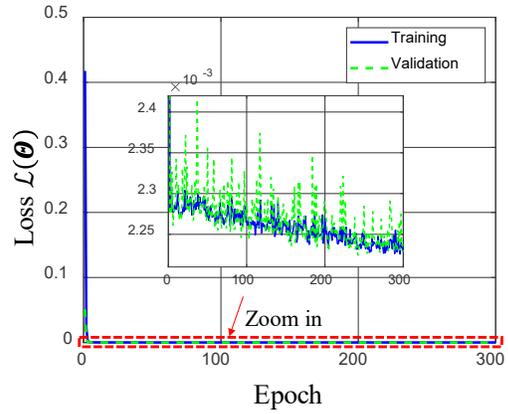

(a) Displacement                  (b) Acceleration

Figure 16. Training and validation loss histories during training process of
the FC dense model in Case-3: (a) Displacement; (b) Acceleration.

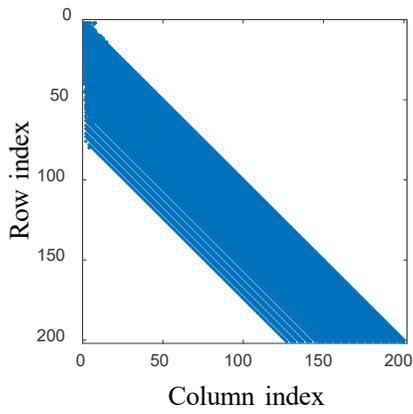 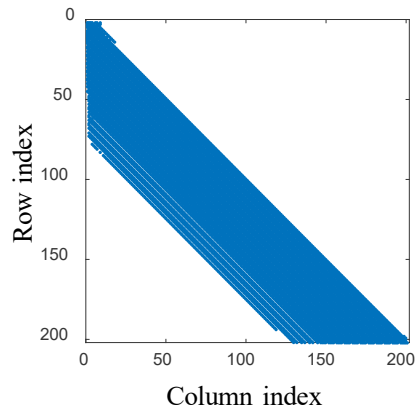

(a) Displacement (#non-zero weights = 8080)     (b) Acceleration (#non-zero weights = 9122)

Figure 17. Sparsity patterns of the weight matrix of (single) FC layer in
the dense models in Case-3: (a) Displacement; (b). Acceleration.

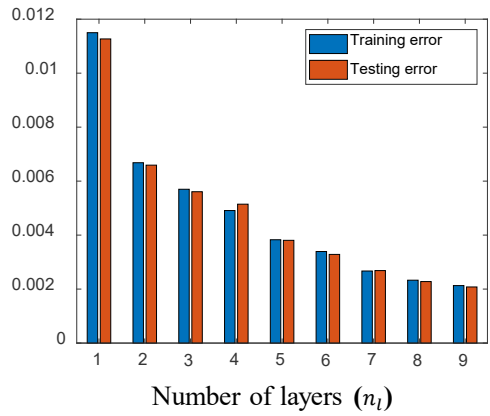

(a) Displacement

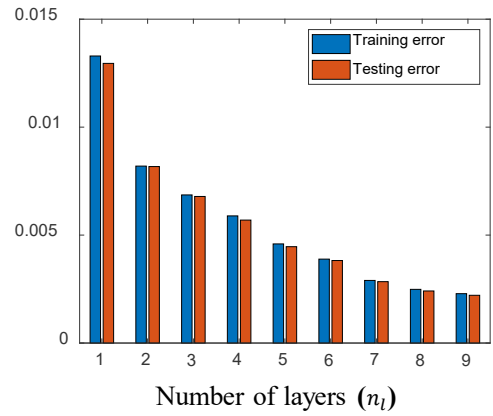

(b) Acceleration

Figure 18. Parametric studies on the optimal number ($n_l$) of CONV layers in the sparse networks in Case-3: (a). Displacement; (b) Acceleration.

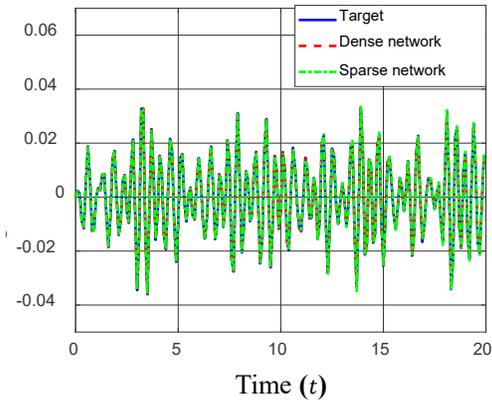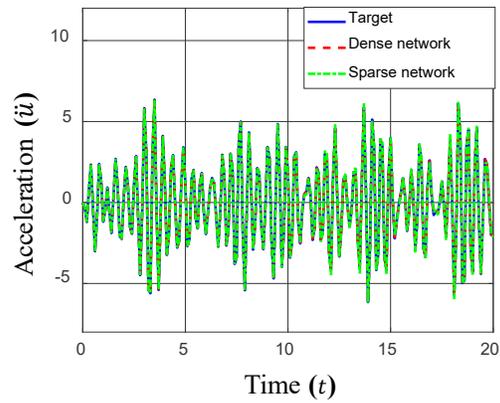

(a) Sample 1

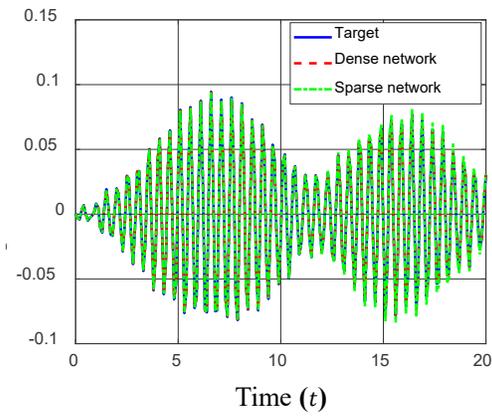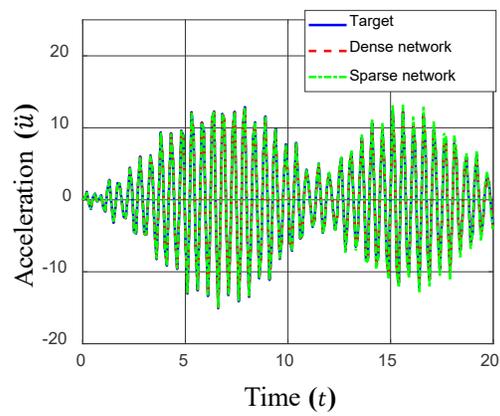

(b) Sample 2

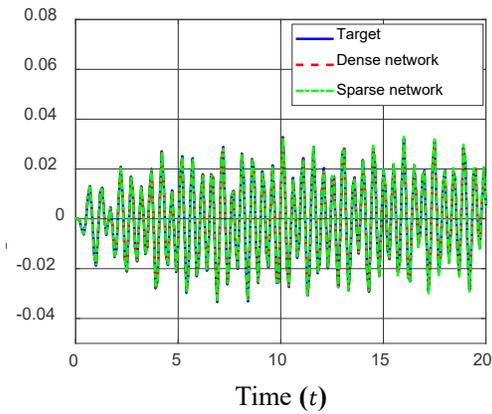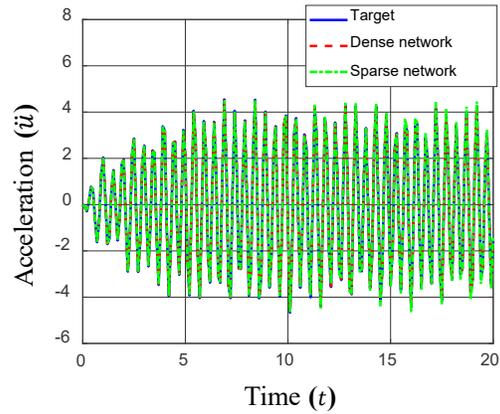

(c) Sample 3

Figure 19. Performance comparison of the dense and sparse network models on three random samples from testing dataset in Case-3: Left – displacement; Right – acceleration.

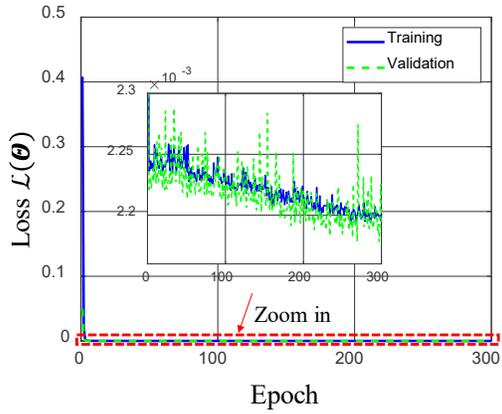 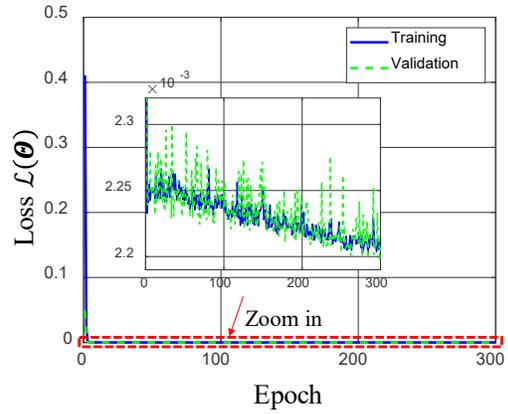

(a) Displacement                    (b) Acceleration

Figure 20. Training and validation loss histories during training process of
the FC dense model in Case-4: (a) Displacement; (b) Acceleration.

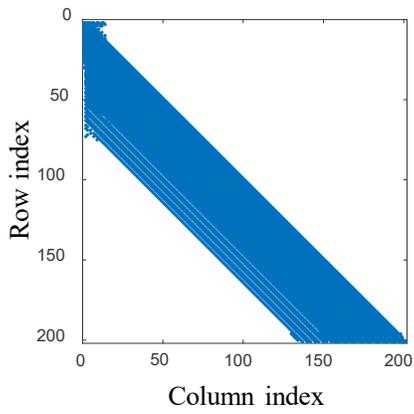 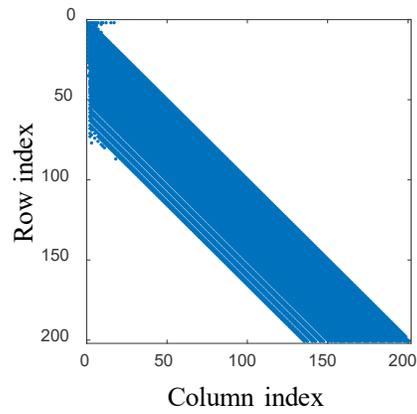

(a) Displacement (#non-zero weights = 8042)    (b) Acceleration (#non-zero weights = 8097)

Figure 21. Sparsity patterns of the weight matrix of (single) FC layer in
the dense models in Case-4: (a) Displacement; (b) Acceleration.

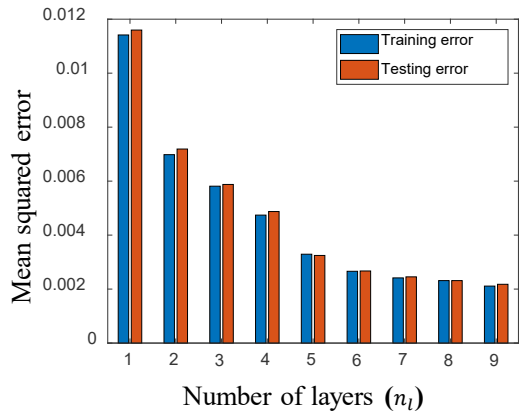 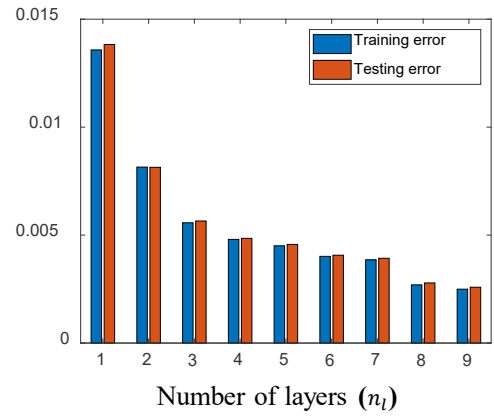

(a) Displacement (b) Acceleration

Figure 22. Parametric studies on the optimal number ($n_l$) of CONV layers
in the sparse networks in Case-4: (a) Displacement; (b) Acceleration.

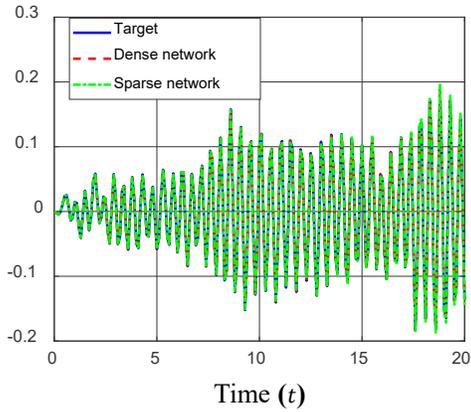
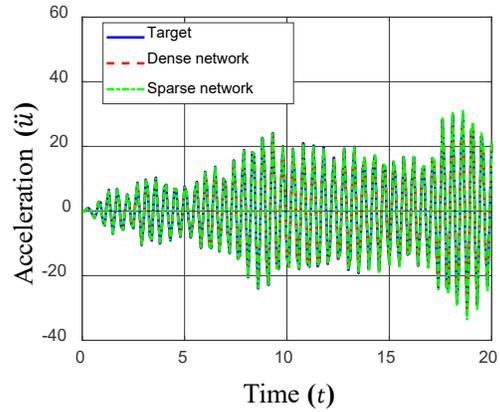

(a) Sample 1

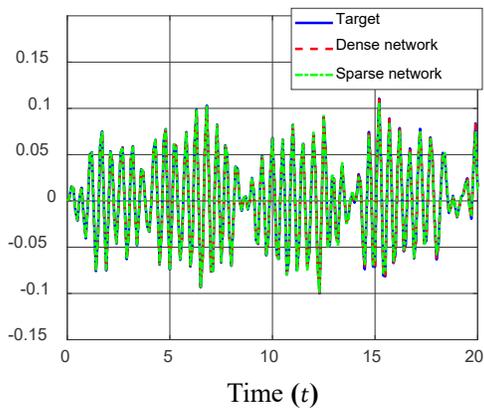
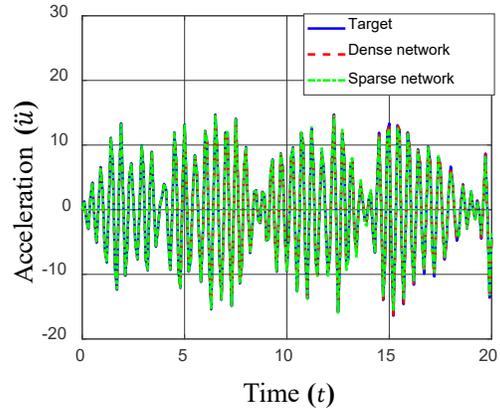

(b) Sample 2

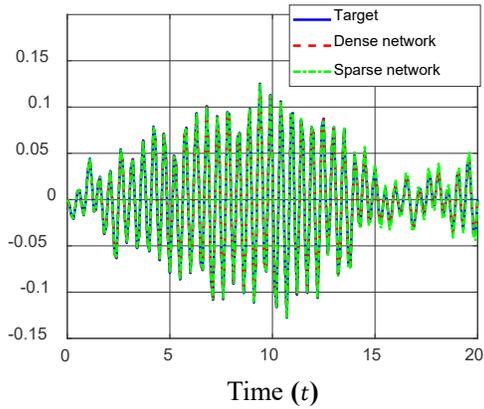
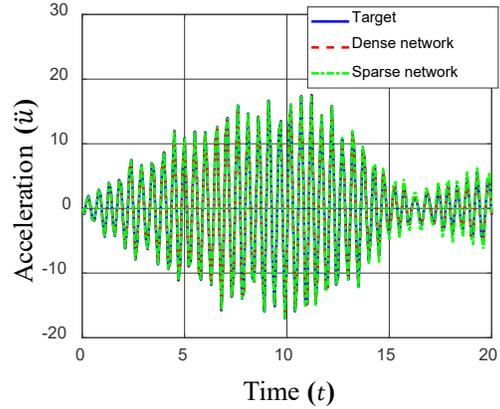

(c) Sample 3

Figure 23. Performance comparison of the dense and sparse network
models on three random samples from testing dataset in Case-4: Left –
displacement; Right – acceleration.

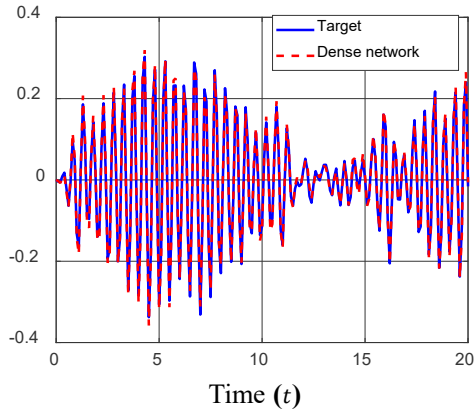 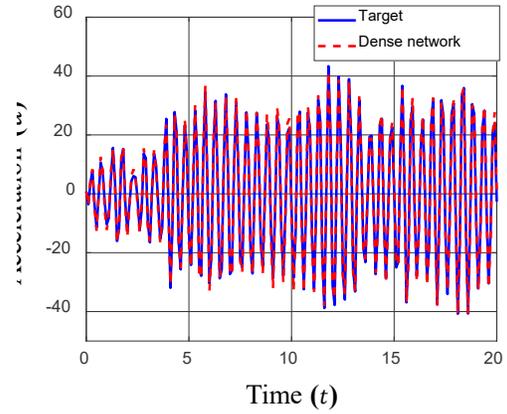

(a) Sample 1

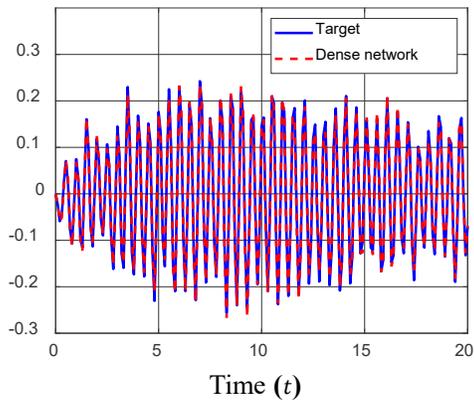 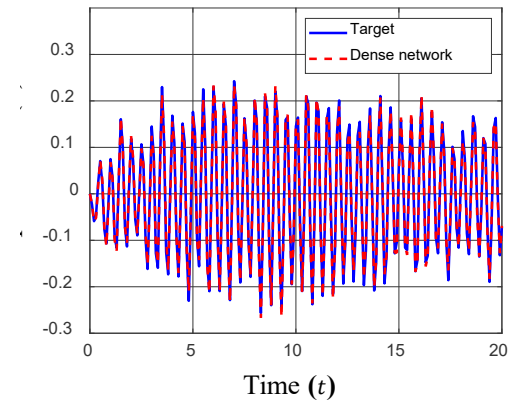

(b) Sample 2

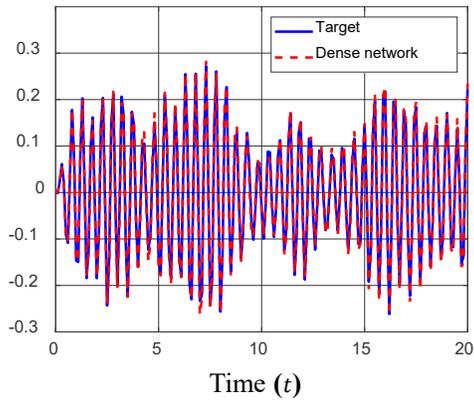 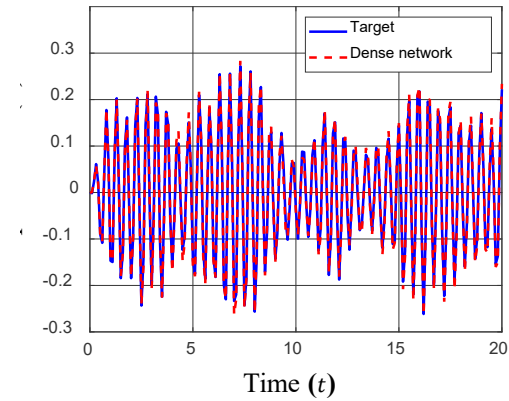

(c) Sample 3

Figure 24. Prediction by the dense network of three random samples from testing dataset in Case-5 with cubic coefficient $b = 0.25k$: Left – displacement; Right – acceleration.

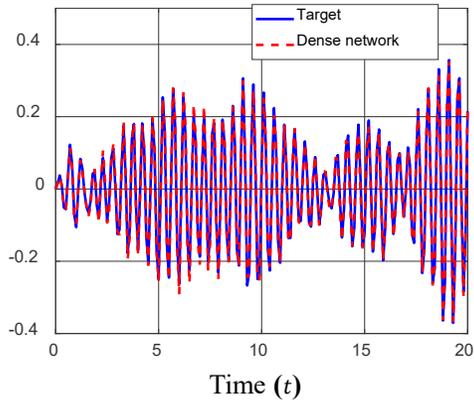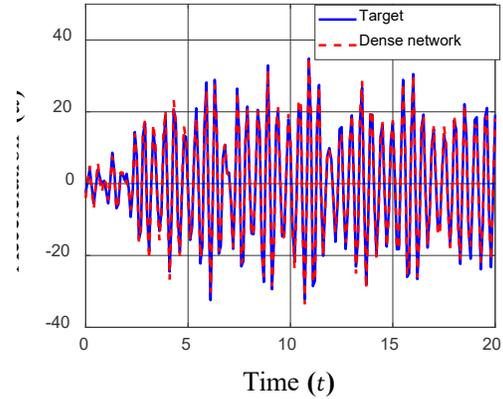

(a) Sample 1

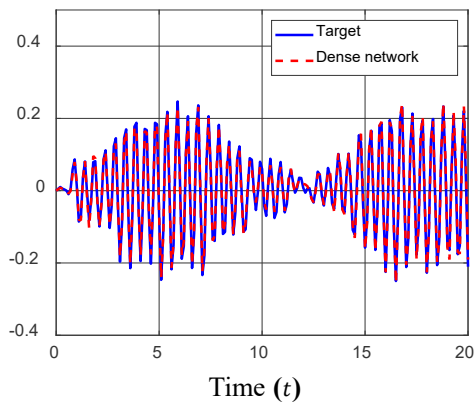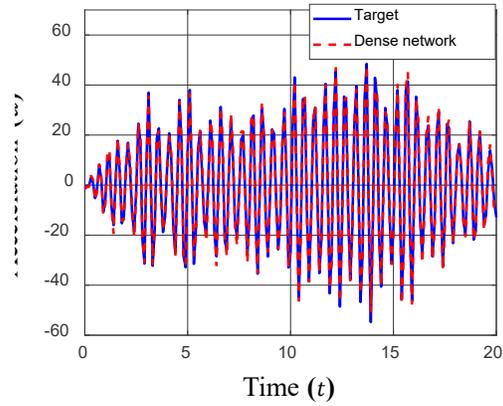

(b) Sample 2

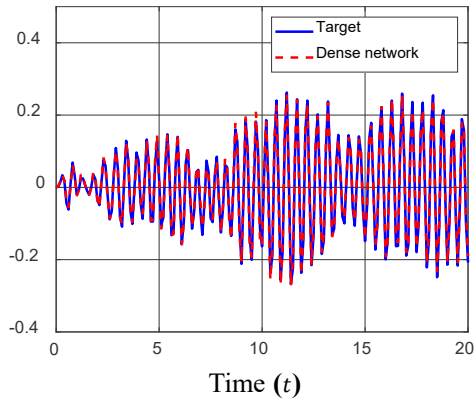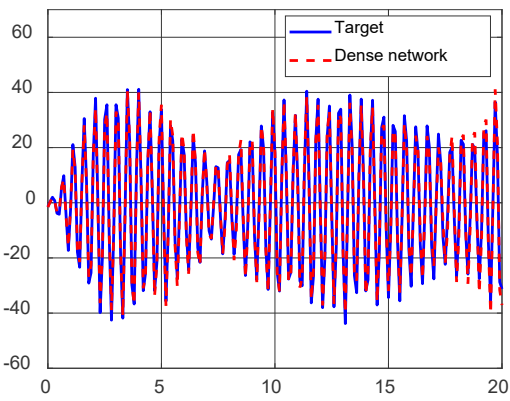

(c) Sample 3

Figure 25. Prediction by the dense network of three random samples from testing dataset in Case-5 with cubic coefficient $b = 0.5k$: Left – displacement; Right – acceleration.

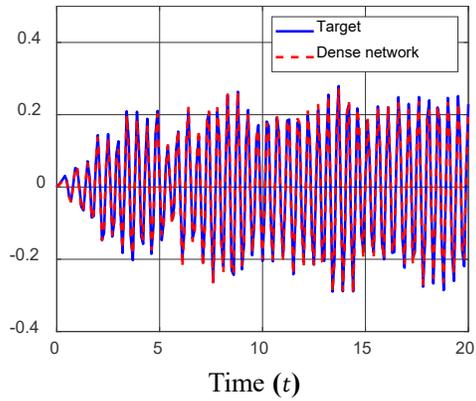 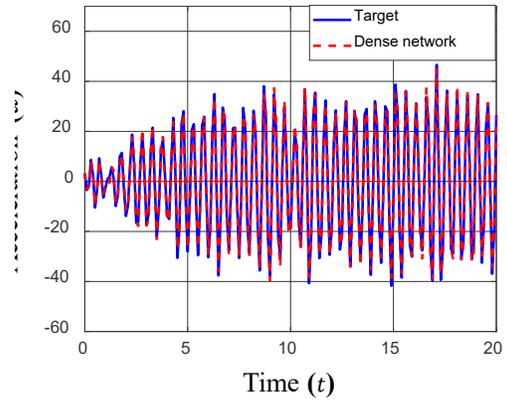

(a) Sample 1

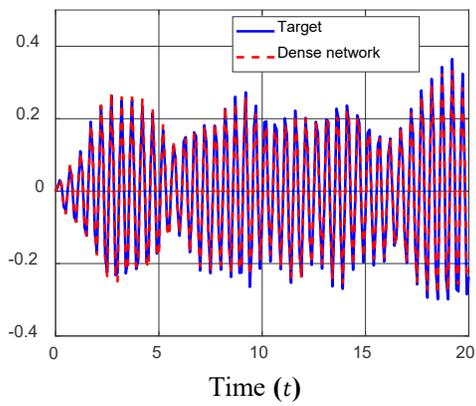 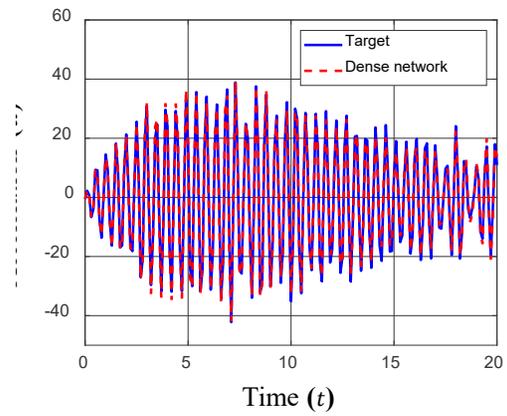

(b) Sample 2

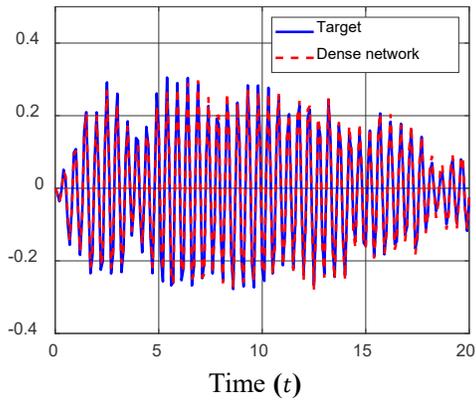 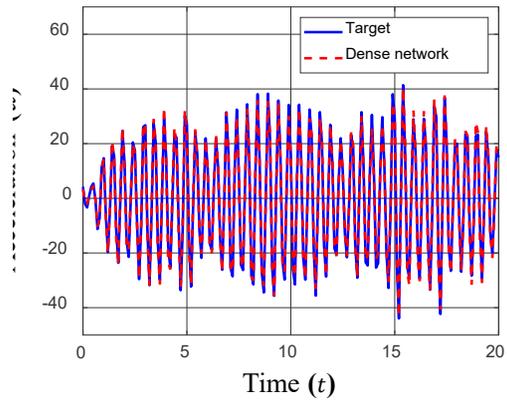

(c) Sample 3

Figure 26. Prediction by the dense network of three random samples from testing dataset in Case-5 with cubic coefficient $b = 0.75k$: Left – displacement; Right – acceleration.

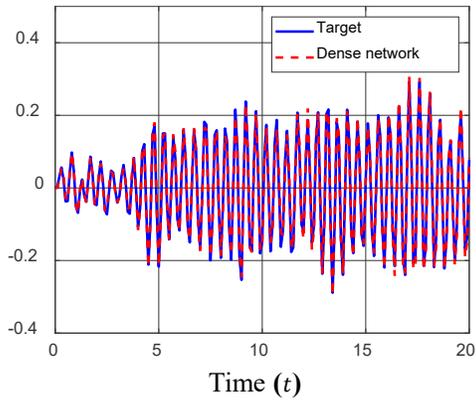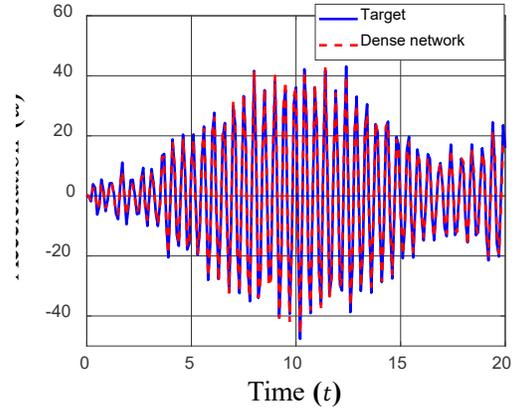

(a) Sample 1

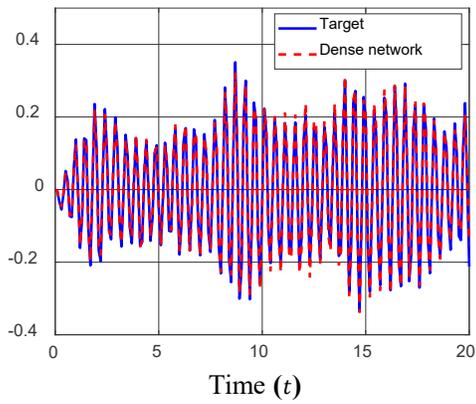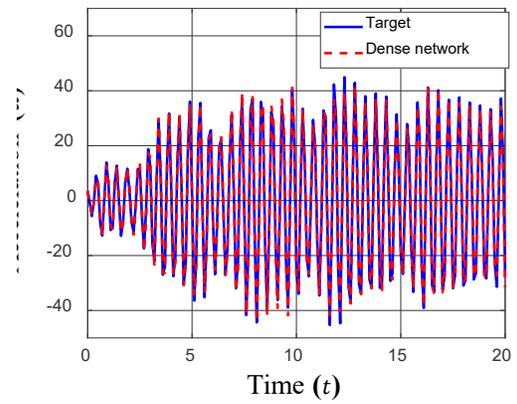

(b) Sample 2

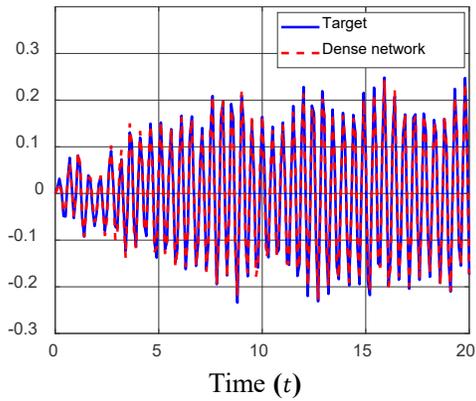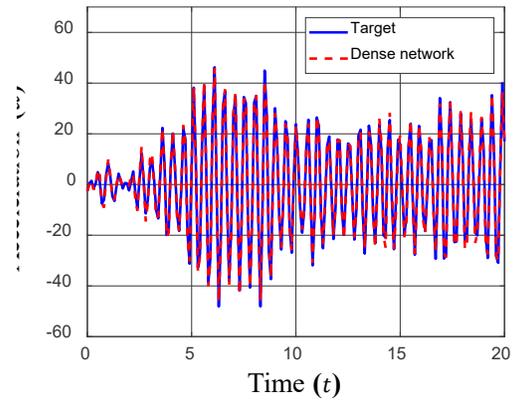

(c) Sample 3

Figure 27. Prediction by the dense network of three random samples from testing dataset in Case-5 with cubic coefficient $b = k$: Left – displacement; Right – acceleration.

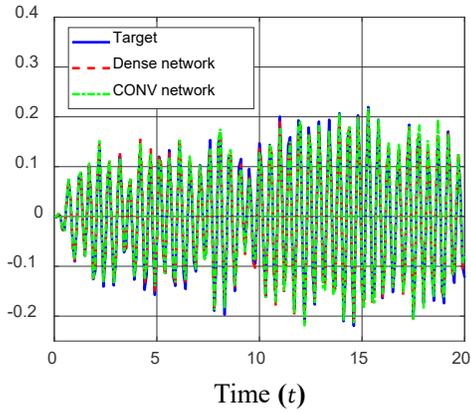
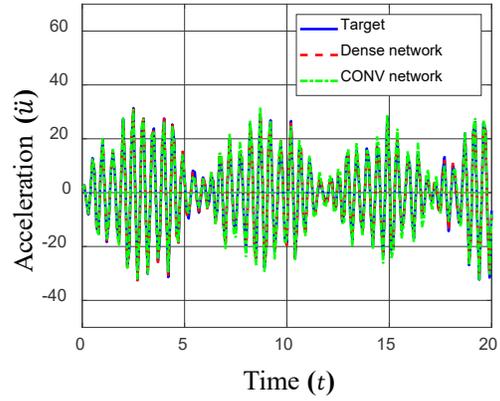

(a) Sample 1

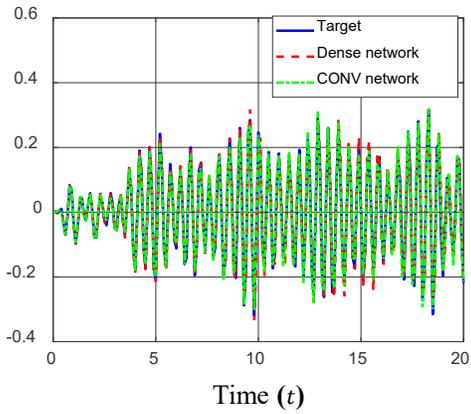
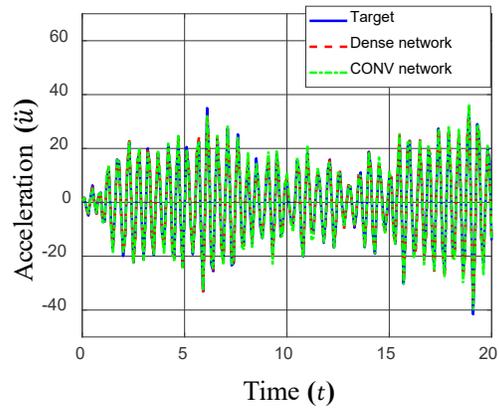

(b) Sample 2

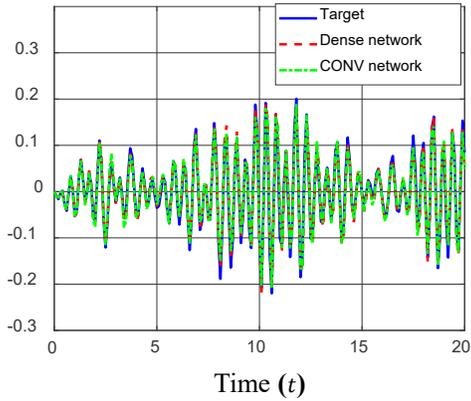
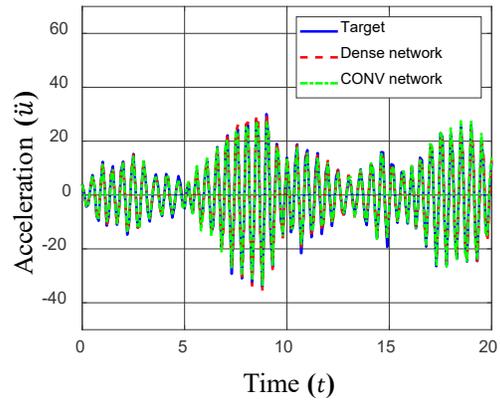

(c) Sample 3

Figure 28. Performance comparison of the CONV enriched NN model and the dense NN model on three random samples from testing dataset in Case-5 with cubic coefficient $b = k$: Left – displacement; Right – acceleration.

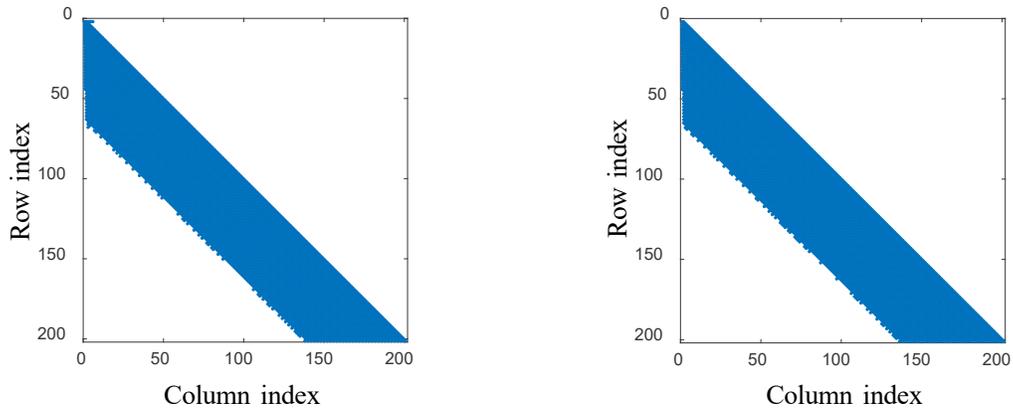

(a) Displacement (#non-zero weights = 5523)   (b) Acceleration (#non-zero weights = 5661)

Figure 29. Sparsity patterns of the weight matrix of (single) FC layer in the dense models for linear MDOF (Case-6) in the loading frequency range $[0,10\pi]$: (a) Displacement; (b) Acceleration.

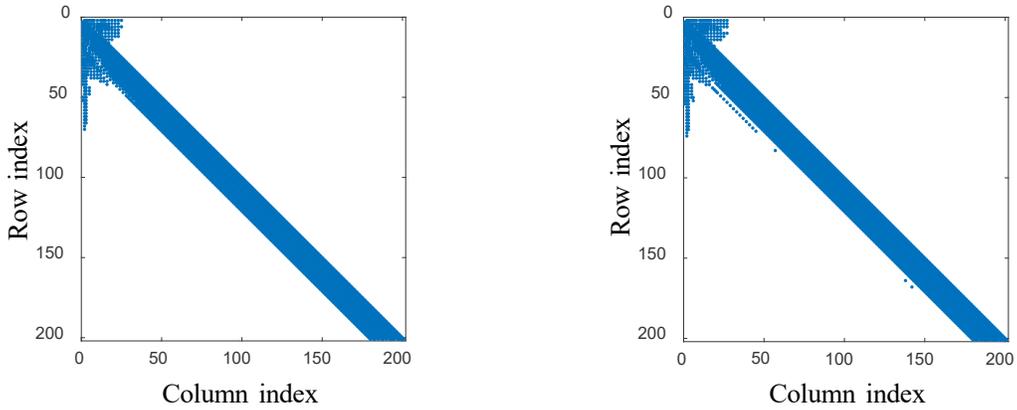

(a) Displacement (#non-zero weights = 3227)     (b) Acceleration (#non-zero weights = 3367)

Figure 30. Sparsity patterns of the weight matrix of (single) FC layer in the dense models for linear MDOF (Case-6) in the loading frequency range $[10,15\pi]$: (a) Displacement; (b) Acceleration.

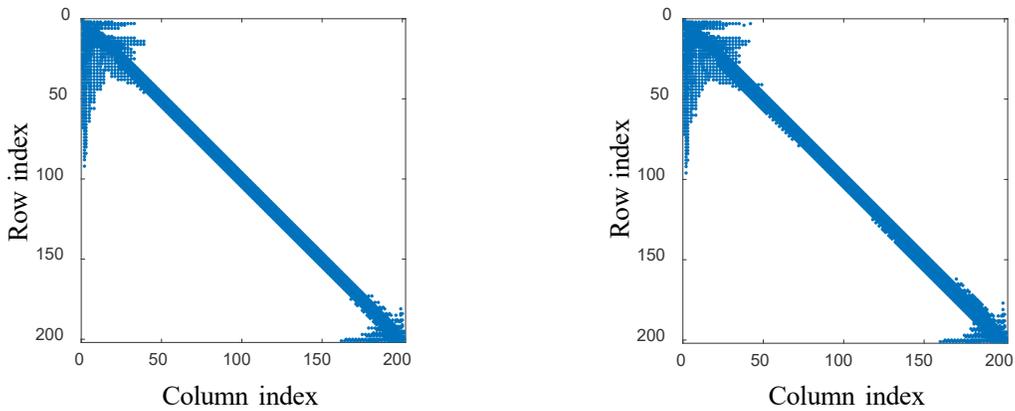

(a) Displacement (#non-zero weights = 2030)     (b) Acceleration (#non-zero weights = 2247)

Figure 31. Sparsity patterns of the weight matrix of (single) FC layer in the dense models for linear MDOF (Case-6) in the loading frequency range $[15\pi, 20\pi]$: (a) Displacement; (b) Acceleration.

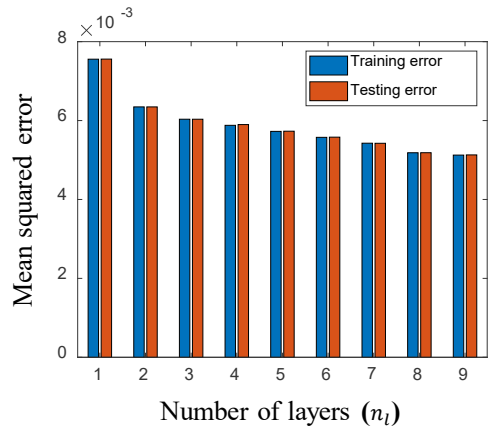
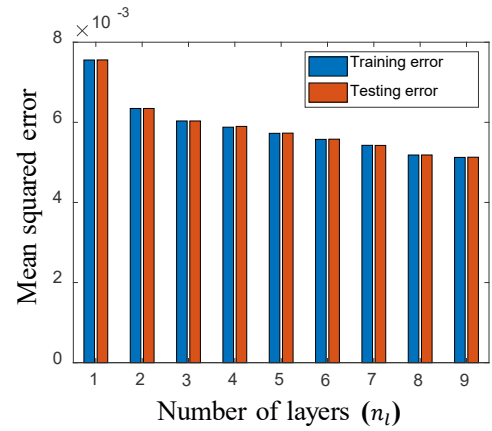

(a) Displacement

(b) Acceleration

Figure 32. Parametric studies on the optimal number ($n_l$) of CONV layers in the sparse networks for linear MDOF (Case-6) in the loading frequency range $[0,10\pi]$: (a) Displacement; (b) Acceleration.

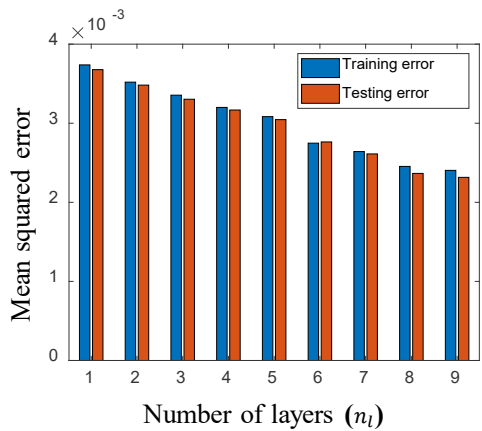
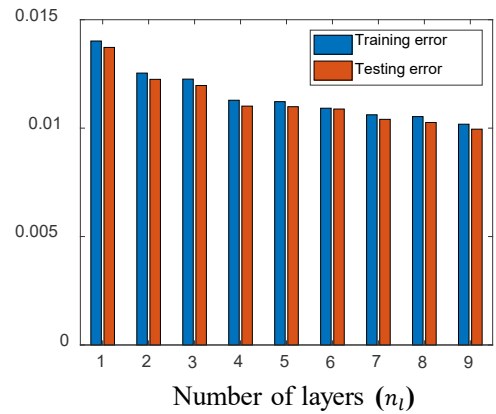

(a) Displacement

(b) Acceleration

Figure 33. Parametric studies on the optimal number ($n_l$) of CONV layers in the sparse networks for linear MDOF (Case-6) in the loading frequency range $[10\pi, 15\pi]$: (a) Displacement; (b) Acceleration.

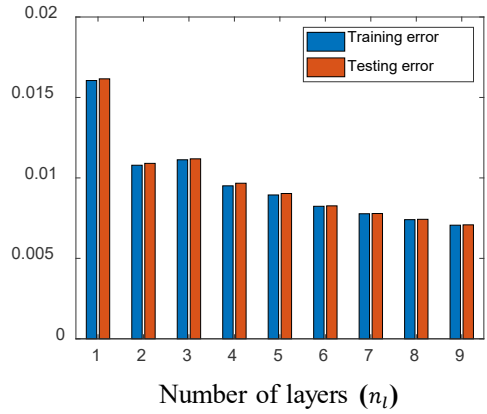

(a) Displacement

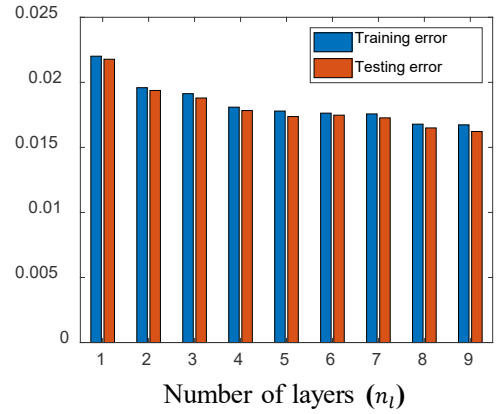

(b) Acceleration

Figure 34. Parametric studies on the optimal number ($n_l$) of CONV layers in the sparse networks for linear MDOF (Case-6) in the loading frequency range $[15\pi, 20\pi]$: (a) Displacement; (b) Acceleration.

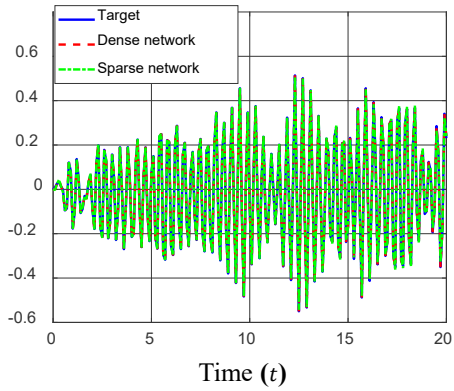 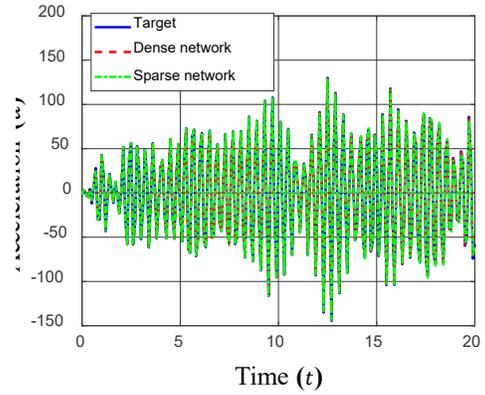

(a) Sample 1

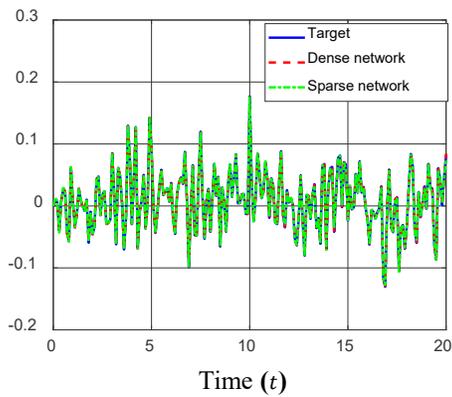 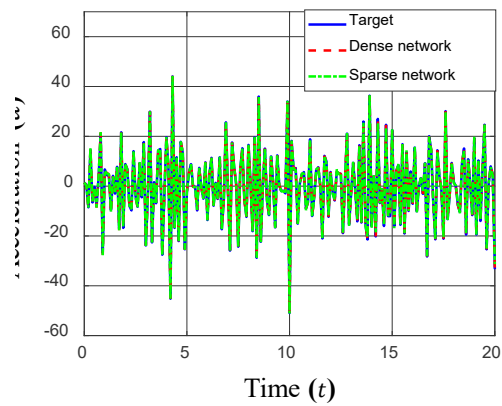

(b) Sample 2

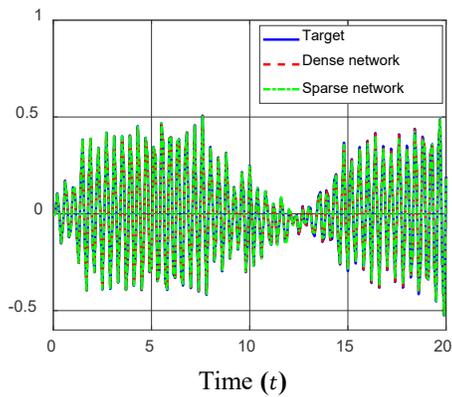 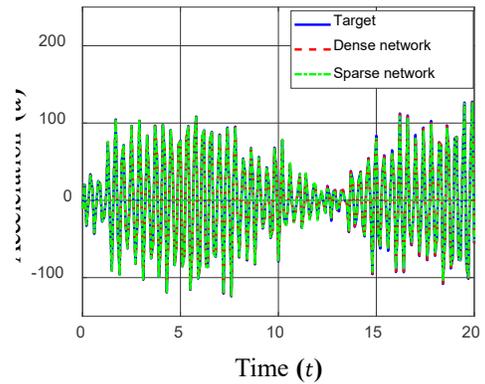

(c) Sample 3

Figure 35. Performance comparison of the dense and sparse network models on three random samples from testing dataset for linear MDOF (Case-6) in the loading frequency range [0,10π]: Left – displacement; Right – acceleration.

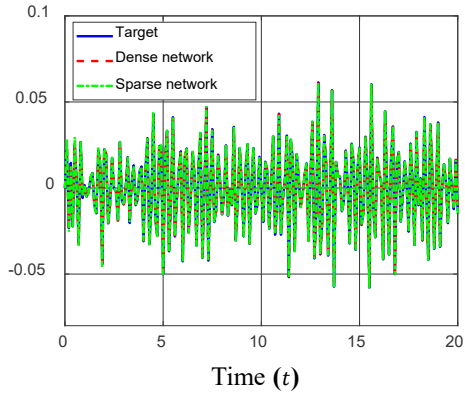
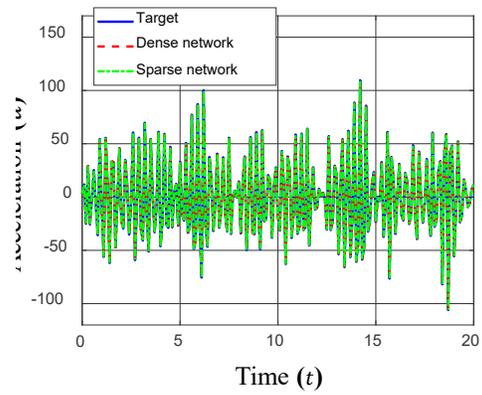

(a) Sample 1

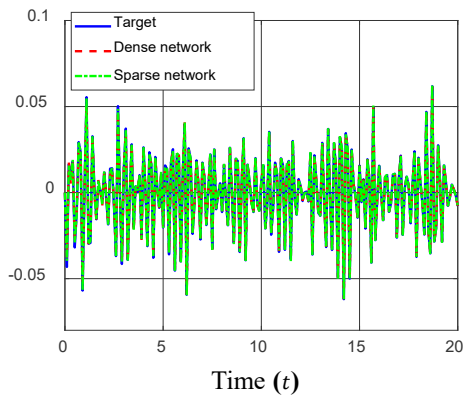
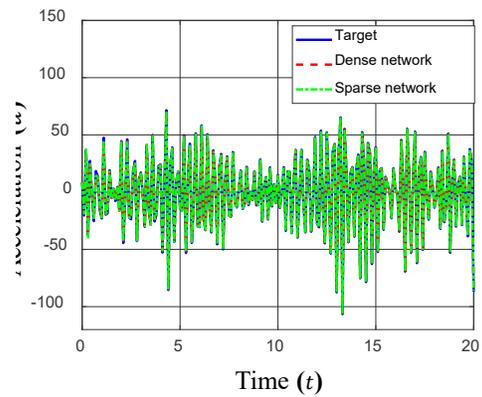

(b) Sample 2

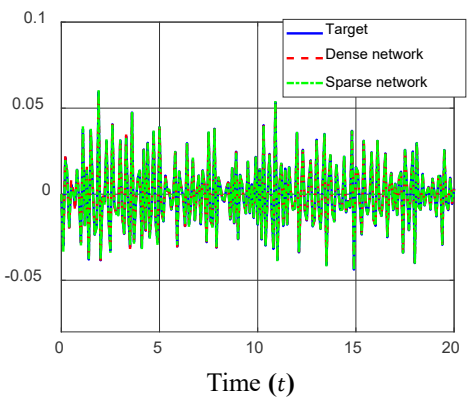
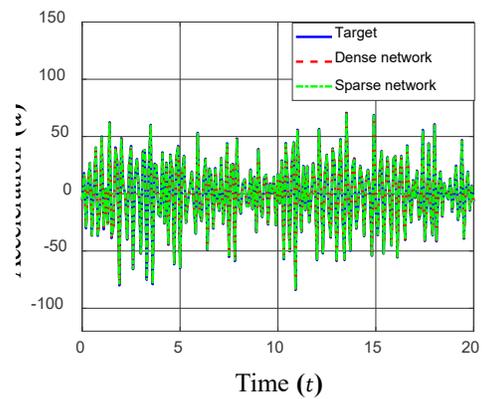

(c) Sample 3

Figure 36. Performance comparison of the dense and sparse network models on three random samples from testing dataset for linear MDOF (Case-6) in the loading frequency range $[10\pi, 15\pi]$: Left – displacement; Right – acceleration.

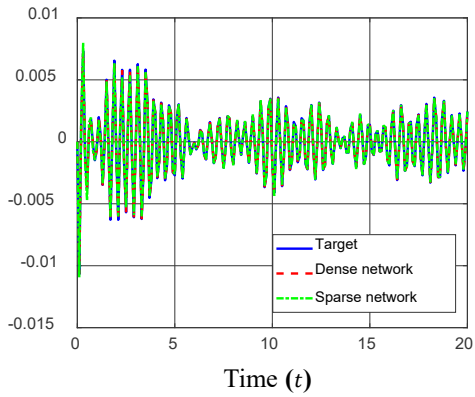 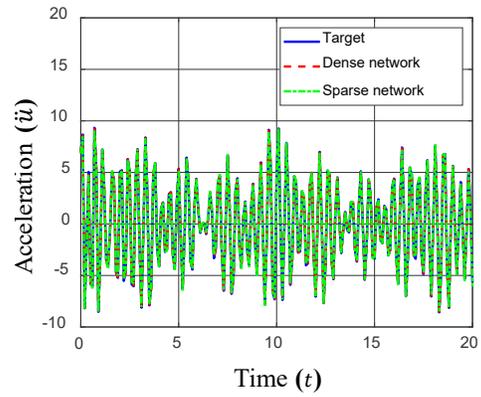

(a) Sample 1

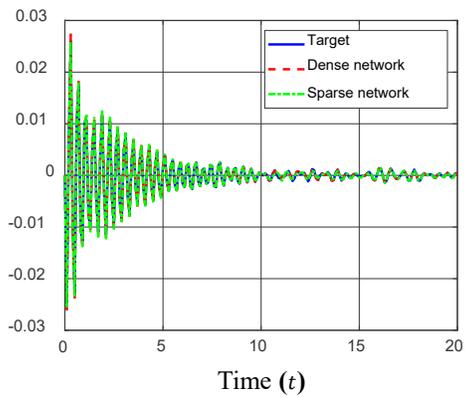 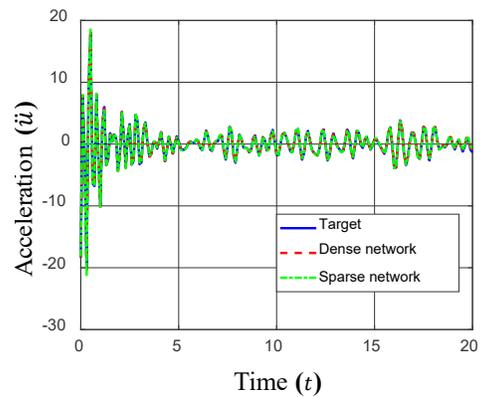

(b) Sample 2

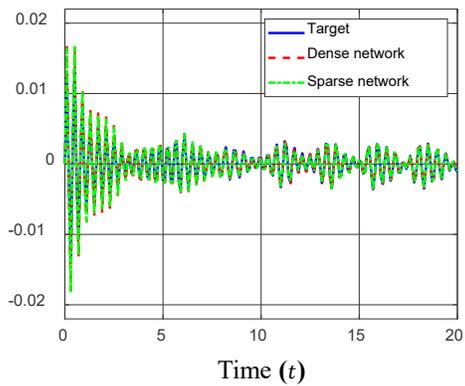 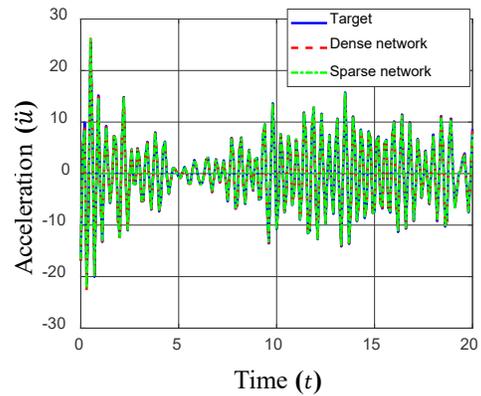

(c) Sample 3

Figure 37. Performance comparison of the dense and sparse network models on three random samples from testing dataset for linear MDOF (Case-6) in the loading frequency range $[15\pi, 20\pi]$: Left – displacement; Right – acceleration.

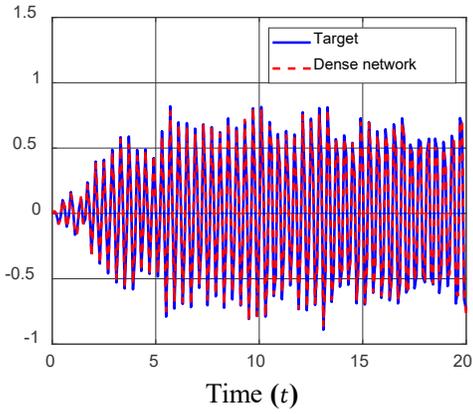 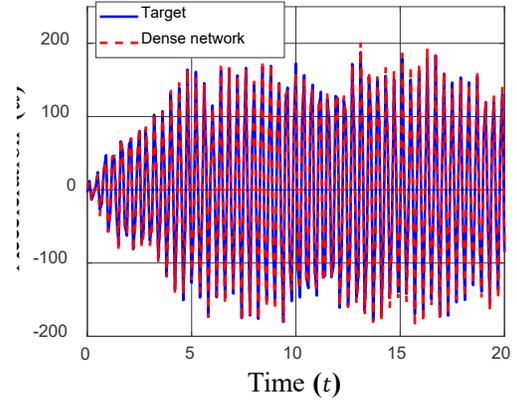

(a) Sample 1

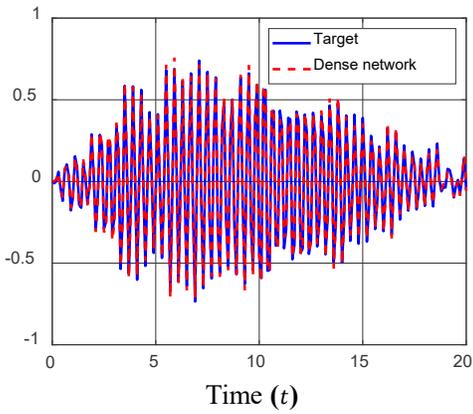 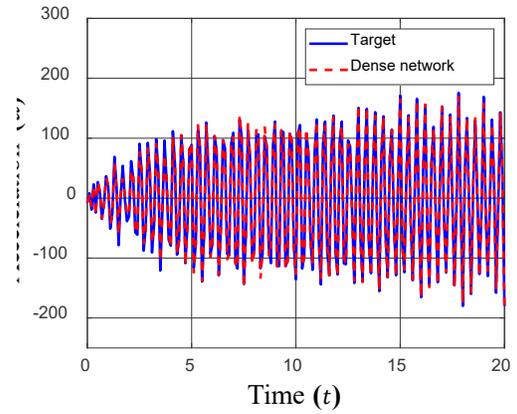

(b) Sample 2

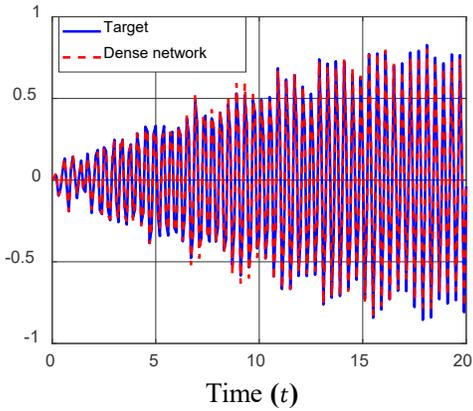 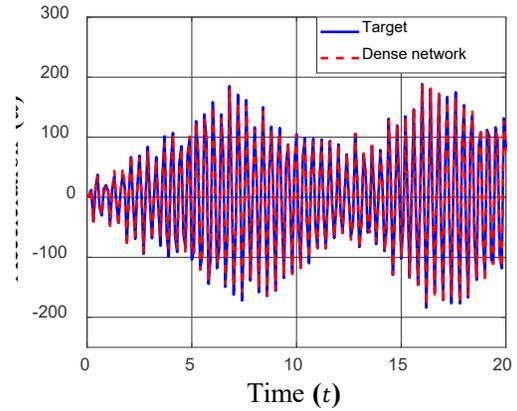

(c) Sample 3

Figure 38. Prediction by the dense network of three random samples from testing dataset in Case-7 with cubic coefficient $b = 0.25k$: Left – displacement; Right – acceleration.

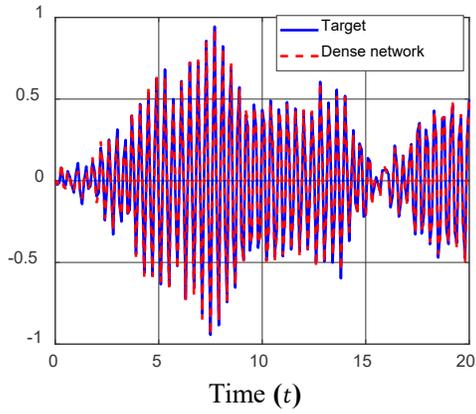 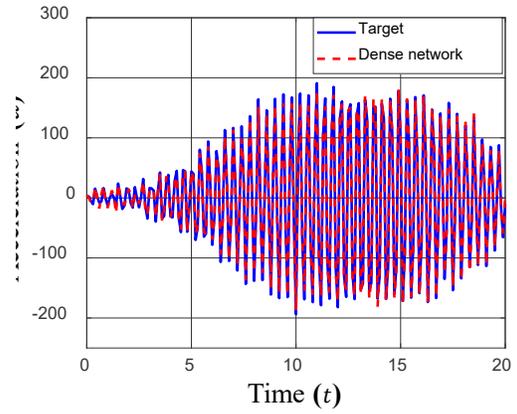

(a) Sample 1

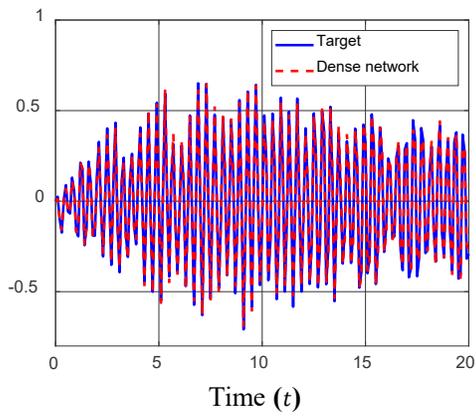 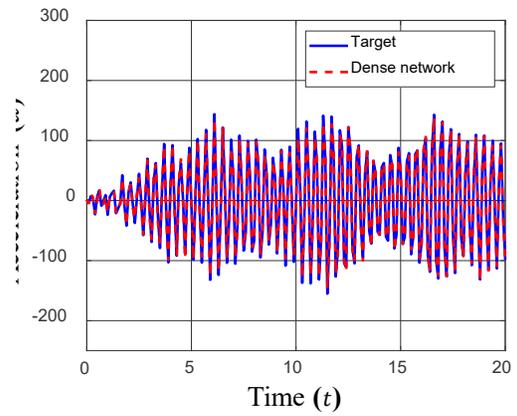

(b) Sample 2

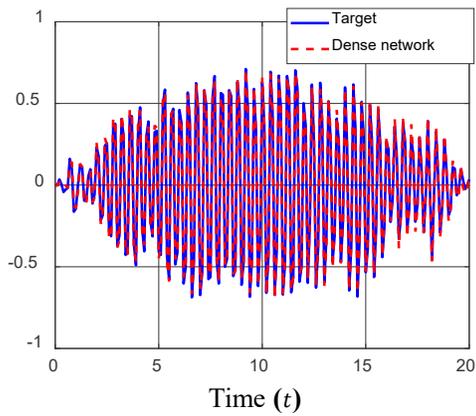 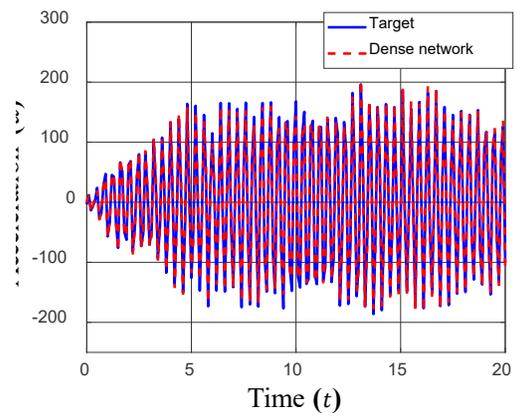

(c) Sample 3

Figure 39. Prediction by the dense network of three random samples from testing dataset in Case-7 with cubic coefficient $b = 0.5k$: Left – displacement; Right – acceleration.

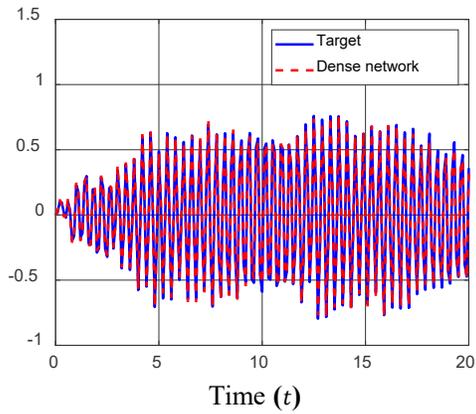 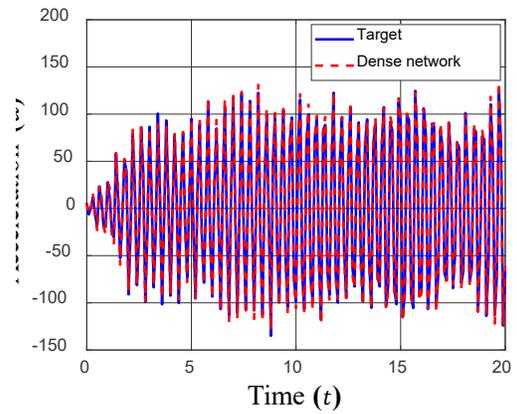

(a) Sample 1

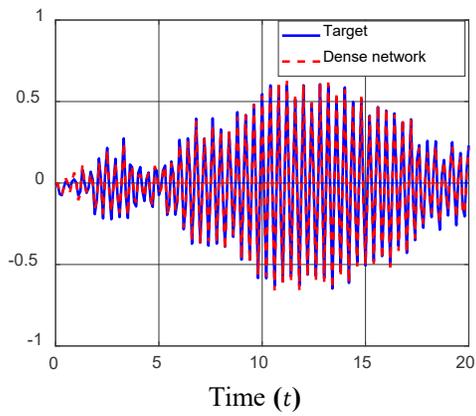 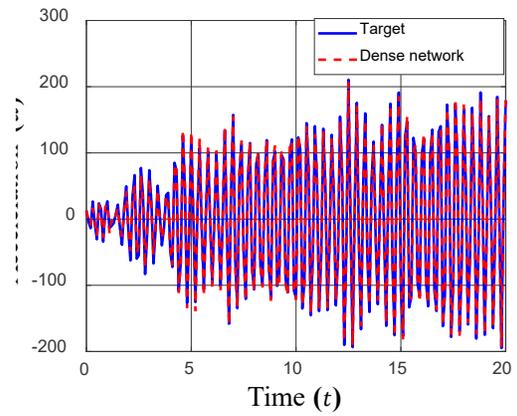

(b) Sample 2

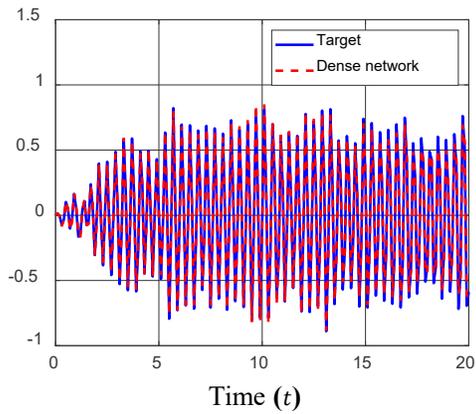 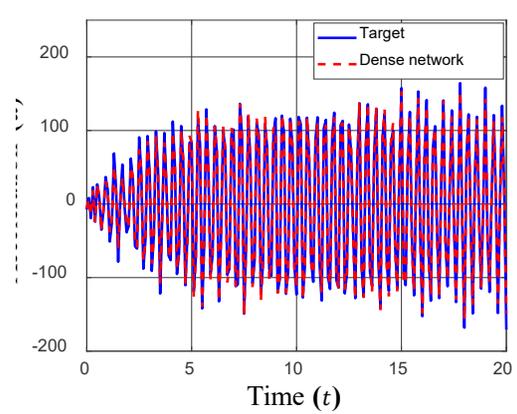

(c) Sample 3

Figure 40. Prediction by the dense network of three random samples from testing dataset in Case-7 with cubic coefficient $b = 0.75k$: Left – displacement; Right – acceleration.

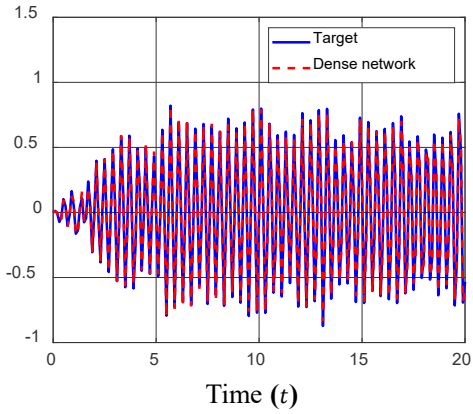 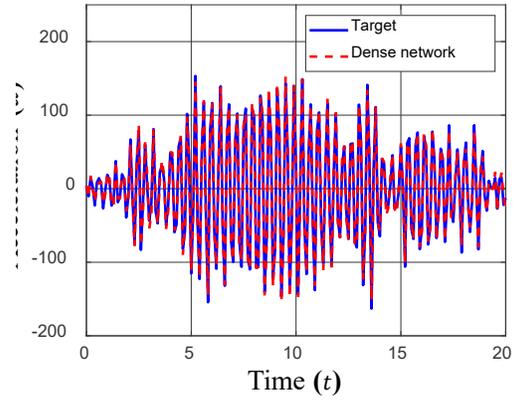

(a) Sample 1

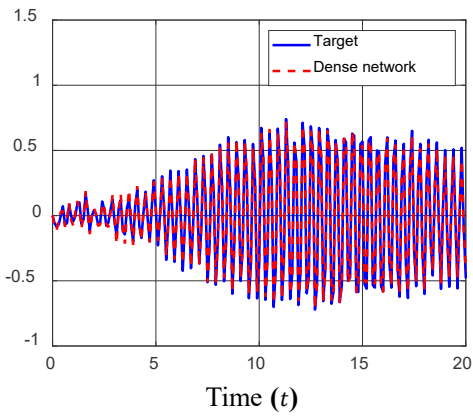 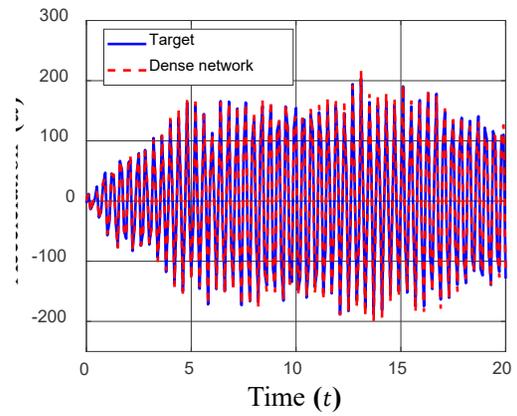

(b) Sample 2

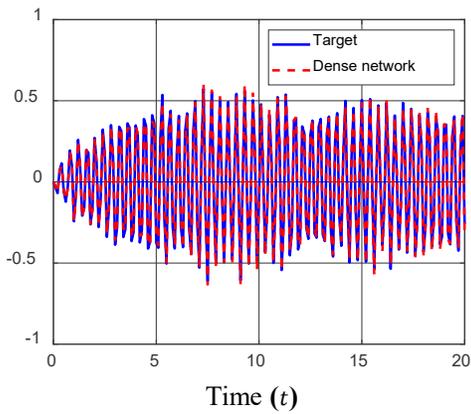 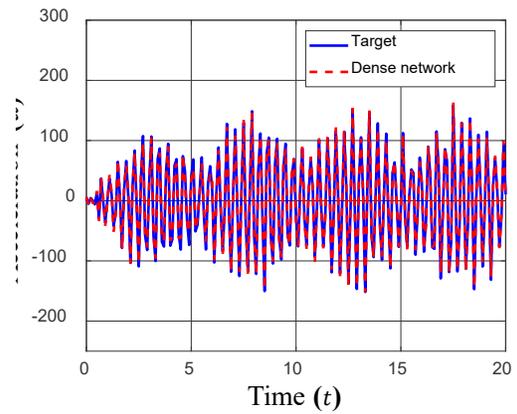

(c) Sample 3

Figure 41. Prediction by the dense network of three random samples from testing dataset in Case-7 with cubic coefficient $b = k$: Left – displacement; Right – acceleration.

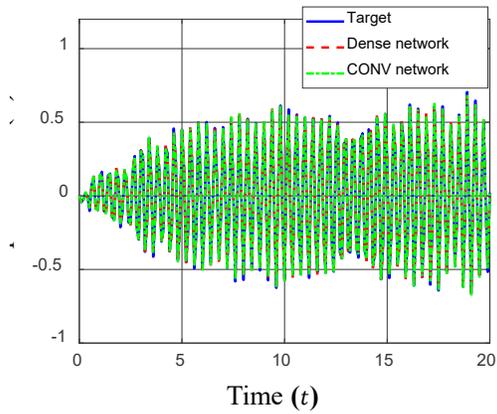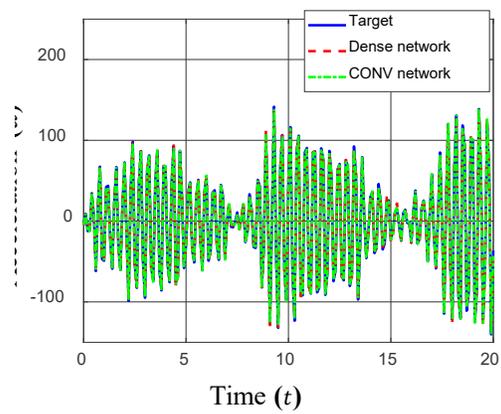

(a) Sample 1

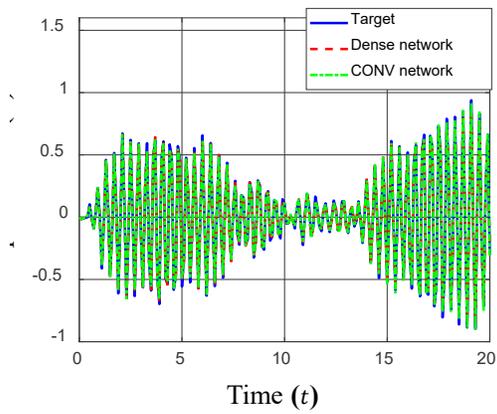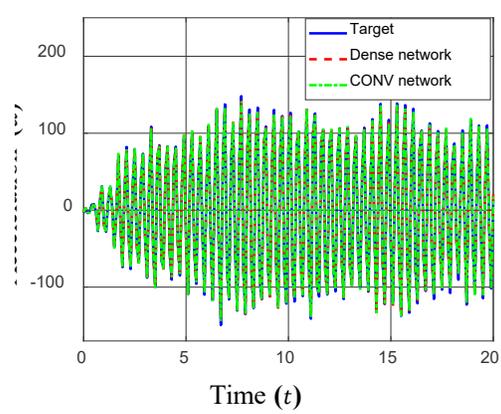

(b) Sample 2

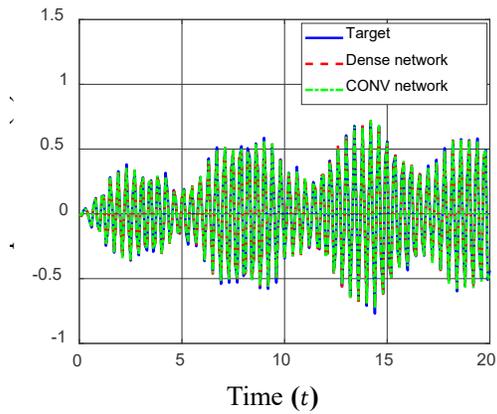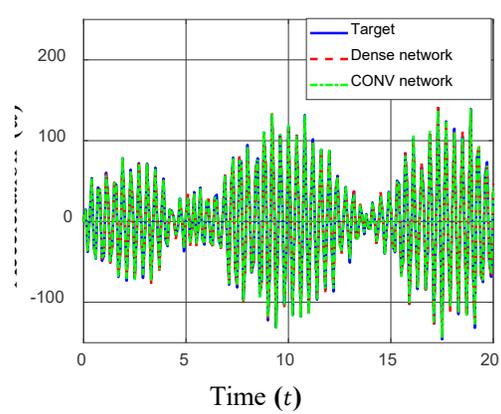

(c) Sample 3

Figure 42. Performance comparison of the CONV enriched NN model and the dense NN model on three random samples from testing dataset in Case-7 with cubic coefficient $b = k$: Left – displacement; Right – acceleration.